\documentclass{article}

\usepackage[preprint]{corl_2026} 

\usepackage{algorithm}
\usepackage{algpseudocode}
\usepackage{float}   
\usepackage{hyperref}       
\usepackage{url}            
\usepackage{booktabs}       
\usepackage{amsfonts}       
\usepackage{nicefrac}       
\usepackage{microtype}      
\usepackage{xcolor}         
\usepackage{enumitem}
\usepackage{graphicx}
\usepackage{subcaption}
\usepackage{amsmath}

\newcommand{\ours}{\text{Action Map Policy}}
\usepackage{caption}
\usepackage{titlesec}
\titlespacing*{\section}
{0pt}{0.2em}{0.1em}

\titlespacing*{\subsection}
{0pt}{0.04em}{0.025em}

\usepackage{graphicx}
\usepackage{wrapfig}
\usepackage{xcolor}
\usepackage{hyperref}
\definecolor{projectblue}{RGB}{225,240,255}

\title{Action Map Policy: Learning 3D Closed-loop Manipulation via Pixel Classification}

\author{
  \textbf{Haojie Huang}$^{1,\ast}$ \quad
  \textbf{Zhang Ye}$^{1}$ \quad
  \textbf{Linfeng Zhao}$^{2}$ \quad
  \textbf{Boce Hu}$^{1}$ \quad
  \textbf{Mingxi Jia}$^{3}$ \\[-0.05em]
  \textbf{Yu Qi}$^{1}$ \quad
  \textbf{Ahmed Agha}$^{1}$ \quad
  \textbf{Dian Wang}$^{2}$ \quad
  \textbf{Robert Platt}$^{1,\dagger}$ \quad
  \textbf{Robin Walters}$^{1,\dagger}$ \\[-0.05em]
  $^{1}$Northeastern University
  \quad
  $^{2}$Stanford University
  \quad
  $^{3}$Brown University \\[-0.05em]
  $^{\ast}$Corresponding author
  \quad
  $^{\dagger}$Equal advising \\[-0.05em]
  \href{https://haojhuang.github.io/amp_page/}{\texttt{haojhuang.github.io/amp\_page}}
  \hfill
  \texttt{huang.haoj@northeastern.edu}
}

\begin{document}
\maketitle


\begin{abstract}
The action space poses a major challenge in robot learning, since it is often high-dimensional, can span long time horizons, and frequently admits multi-modal optimal solutions.
A good choice of action representation and loss function can help to address these concerns, but there are often trade offs.
We propose Action Map Policy (AMP), which casts 3D closed-loop manipulation policy learning as a classification problem in image space. 
While classification has been an effective formulation in generative language models, applying it to robot action learning is difficult because naively discretizing high-dimensional continuous actions explodes the token vocabulary.
Our key idea is to project 3D actions onto the camera image planes and treat each pixel location as a discrete class, thus controlling dimensionality while retaining multi-modality. 
This method supports millimeter-level precision for high-dimensional actions without requiring a prohibitively large vocabulary, while preserving fine-grained pixel-wise visual signals.
Furthermore, it can predict the entire action chunk in a single forward pass, avoiding complex noise scheduling and iterative denoising while achieving substantially faster inference than diffusion policies.
Experiments on various manipulation tasks show that AMP outperforms strong baselines, achieving higher success rates, faster inference, and enhanced spatial reasoning.
\end{abstract}
\section{Introduction}
Modern large language models (LLMs)~\cite{achiam2023gpt,team2024gemini,touvron2023llama} have achieved remarkable success by casting sequence prediction as classification, delivering both strong predictive accuracy and powerful multi-modal distribution modeling. Extending this paradigm to robot action learning is appealing, since action distributions, like language, are inherently multi-modal and often admit several optimal solutions for a given state. A naive formulation, however, discretizes each translation and rotation component independently, producing a combinatorial explosion of tokens: with only 10 bins per dimension, a single 6-DoF action already requires $10^6$ classes, and this cost compounds further over the action horizon. As a result, discrete tokenization struggles to faithfully represent high-dimensional continuous actions, failing to capture the geometric structure of large action spaces and degrading performance on fine-grained manipulation tasks ~\cite{goyal2025vla,liu2026oat,zitkovich2023rt,kim2024openvla}.

\begin{figure*}[t]
    \centering
     \setlength{\abovecaptionskip}{2pt}
    \setlength{\belowcaptionskip}{0pt}
    \includegraphics[width=0.9\linewidth]{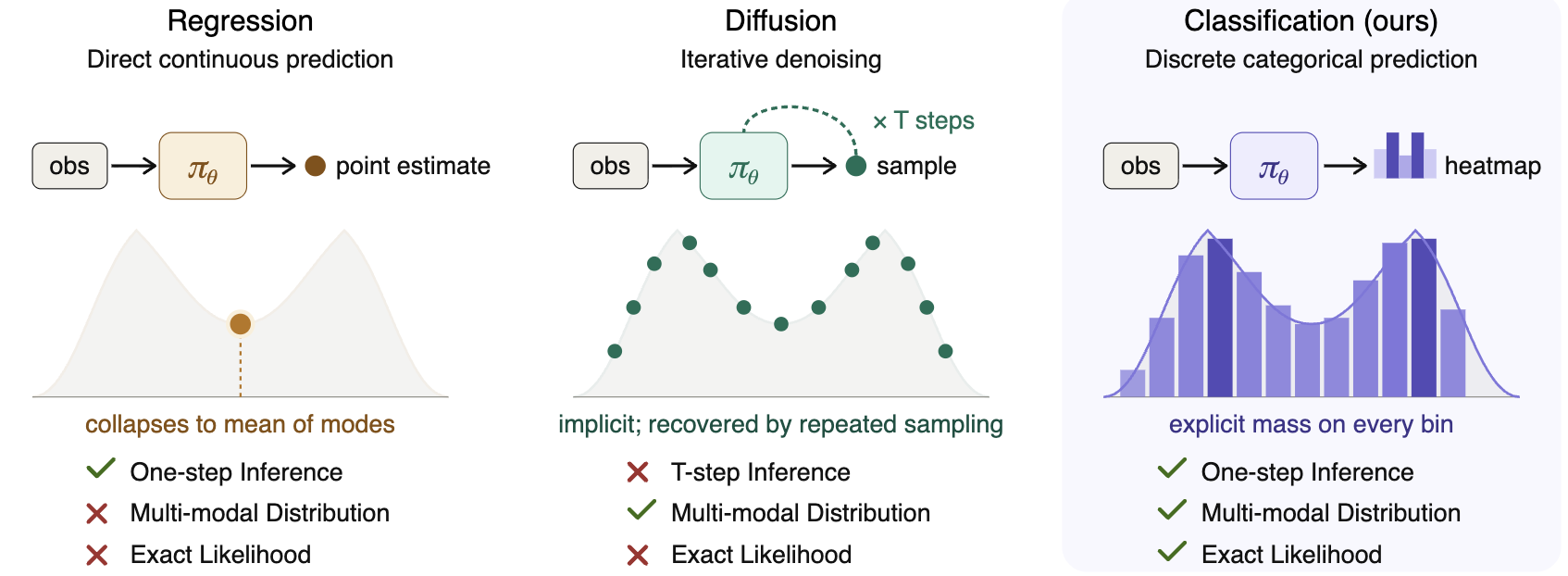}
    \caption{Comparison of different policy-learning frameworks. Regression methods can easily collapse to the mean of multiple modes. Diffusion methods can, in principle, capture multi-modal distributions, but require multiple denoising steps to generate a single sample and sampling-based procedures to recover the full distribution. In contrast, our method supports single-pass inference and captures the full action distribution using a dense map that represents actions geometrically.}
    \label{fig:intro}
\end{figure*}

To address these challenges, we propose \ours{} (AMP), a new formulation that casts 3D closed-loop manipulation policy learning as a classification problem in image space. Instead of regressing high-dimensional continuous actions~\cite{zhao2023learning,haldar2024baku,mandlekar2021matters} or quantizing them into discrete action tokens~\cite{goyal2025vla, liu2026oat,zitkovich2023rt,kim2024openvla}, we represent 3D action learning as a pixel classification task in each 2D camera image plane. Specifically, we define the end-effector pose as a set of 3D keypoints and let the policy predict which pixel each keypoint projects to in every camera view. This turns policy learning into predicting a distribution over pixels for every keypoint across future timesteps.
This representation captures pixel-level precision while avoiding the combinatorial blow-up of per-dimension discretization. Our experiments show that it can provide the millimeter-level precision required for complex 3D tasks, including coffee preparation and toaster insertion. Additionally, it enables manipulation policies to be trained directly with a cross-entropy objective, which naturally models the multi-modal mass distribution over plausible trajectories. Several additional features further distinguish AMP from existing policy learning methods. As summarized in Figure~\ref{fig:intro}, it generates an action chunk in a single forward pass, avoiding complex iterative sampling procedures and enabling much faster inference. Moreover, its output can be interpreted as an explicit action distribution, which may provide a natural interface for future refinement, such as reweighting actions or fine-tuning with reinforcement learning~\cite{peng2019awr,schulman2017proximal}.

To realize AMP with multi-camera inputs, we design a novel backbone architecture that we call \textit{X-Net} that fuses multi-view images and maps them directly to action feature maps, preserving dense spatial structure rather than collapsing the scene into a single global embedding. 
AMP aligns the action representation with the observation, allowing the model to reason more effectively in image space. Because each input pixel is associated with a corresponding action prediction, the policy  remains grounded in the spatial structure of the scene and can exploit fine-grained, pixel-level cues. 
We demonstrate this feature in a simple test with several visually identical objects, where the target is specified only by a laser pointer. Our method reliably responds to this small visual cue, while Diffusion Policy~\cite{chi2025diffusion} largely ignores it. This is important because it enables the policy to use small but task-critical visual cues, allowing the robot to act more precisely and responsively in real-world manipulation.

Our contributions are summarized as follows:
(1) We propose a novel framework for 3D closed-loop manipulation policy learning that represents actions as dense heatmaps over a temporal horizon, reformulating policy learning as a classification problem;
(2) We design a novel backbone that we call \textit{X-Net} that maps multi-view observations to dense action maps, preserving fine-grained visual cues while explicitly modeling spatial action likelihoods;
(3) Experiments demonstrate that our method outperforms strong baselines in both simulated and real-world settings, achieving higher success rates, faster inference, and stronger spatial reasoning ability.



\section{Related Work}
\textbf{Manipulation Action Representation:}
Action representation largely determines how robot policies interact with and learn from the physical world. Existing manipulation policies mainly differ in whether actions are represented as compact low-dimensional commands or as dense spatial structures.
\textbf{Low-dimensional action representations} are a common choice, where actions live in a compact continuous space such as end-effector poses or joint commands. Many policy learning methods directly regress or diffuse these low-dimensional actions~\cite{wang2024equivariant,mandlekar2021matters,hu20263d,chi2025diffusion}. Other works map actions into alternative latent or discrete spaces. For example,~\cite{zhao2023learning} trains a CVAE~\cite{NIPS2015_5775} to encode actions into latent style vectors, while~\cite{liu2026oat,zitkovich2023rt,kim2024openvla} tokenizes continuous actions into discrete action tokens.
\textbf{Dense action representation} has also been explored, especially in open-loop pick-and-place settings. For example,~\cite{zeng2021transporter,huang2022equivariant,jia2025learning} represents pick-and-place actions as dense feature maps, where each pixel corresponds to a spatial location and the channel dimension encodes rotation. This idea has been extended to 3D action spaces in~\cite{shridhar2023perceiver,huang2024fourier,goyal2023rvt}. Related methods also represent actions through dense goal configurations, such as target point clouds to specify desired manipulation outcomes~\cite{pan2023tax, eisner2024deep, huang2024imagination, huang2025match}. In contrast, our method uses dense heatmaps to represent closed-loop 3D manipulation actions over a temporal horizon, enabling policy learning through classification.

\textbf{Manipulation Learning Framework:}
We categorize existing 3D closed-loop manipulation policy learning frameworks into three major classes.
\textbf{Regression-based methods} typically predict continuous actions directly and train the policy with an MSE loss~\cite{zhao2023learning,haldar2024baku,mandlekar2021matters}. While simple and efficient, these methods can struggle to model multi-modal action distributions, since minimizing MSE may collapse multiple valid actions into an averaged prediction~\cite{pearce2023imitating, pan2025much}.
\textbf{Iterative generative policies} include diffusion- and flow-matching-based methods~\cite{chi2025diffusion,hu2024adaflow,lipman2022flow,yan2025maniflow}. These methods model action distributions through iterative sampling procedures, often relying on noise schedules and multiple refinement steps to generate actions. In principle, they can represent multi-modal action distributions by learning the underlying score or flow field. However, because the distribution is represented implicitly through iterative generation, it is difficult to directly inspect, constrain, or supervise the likelihood of specific action candidates.
\textbf{Classification-based methods} often follow the LLM-style autoregressive paradigm~\cite{goyal2025vla,zitkovich2023rt,kim2024openvla}: they discretize continuous actions into action tokens and train the policy to predict the next action token or token chunk. This formulation benefits from a stable cross-entropy objective, but it can be difficult to apply to fine-grained manipulation tasks: nearby continuous actions may be mapped to the same discrete token, while increasing the resolution can lead to a prohibitively large vocabulary. In contrast, our method addresses this limitation by representing actions as pixel-wise heatmaps, enabling fine-grained action prediction through classification without requiring an excessively large action vocabulary.

\textbf{Policy Learning using Keypoints:}
Keypoint learning has been widely studied in computer vision for tasks such as human pose estimation, facial landmark detection, and point tracking. In robot manipulation, keypoints provide a compact and geometrically meaningful way to represent actions~\cite{haldar2026point}. A set of 3D keypoints can jointly encode translation and rotation, since the end-effector pose can be recovered from their spatial configuration. Prior works have used 3D keypoints for pick-and-place manipulation~\cite{huang2022edge,hu2024orbitgrasp,huang2024rekep,manuelli2019kpam} and leveraged the pointing ability of vision-language models with depth observations for manipulation control~\cite{yuan2024robopoint,cheng2025pointarena}.
Recent methods further use keypoints as intermediate action representations for policy learning. For example,~\cite{ren2025motion,xu2024flow} generate 2D keypoint trajectories by diffusing their coordinates. 
Our method builds on the strength of keypoints for structured action prediction, but differs in a key way: rather than predicting low-dimensional keypoint coordinates, it predicts dense keypoint distributions over the image space across a temporal horizon, enabling closed-loop 3D manipulation policy learning through classification.


\section{Method}
\label{sec:method}
\textbf{Setup:}
We focus on 3D closed-loop manipulation policy learning from image  inputs. Given the current observation $\mathcal{O}_t=(o_t^1,o_t^2,\cdots,o_t^n)$ captured from $n$ cameras with known intrinsics and extrinsics, our objective is to learn a policy function $\pi \colon \mathcal{O}_t \mapsto \mathcal{A}_t$ from observations to an action chunk  $\mathcal{A}_t=(a_{t},\cdots,a_{t+l-1})$ which represents a sequence of $l$ actions. Each action specifies an end effector pose   $a=(T,R,w)$, where $T\in \mathbb{R}^3$, $R\in \mathrm{SO}(3)$ and $w\in \mathbb{R}$ denote the translation, rotation and gripper aperture, respectively. 
\begin{figure*}[t]
    \centering
    \includegraphics[width=0.95\linewidth]{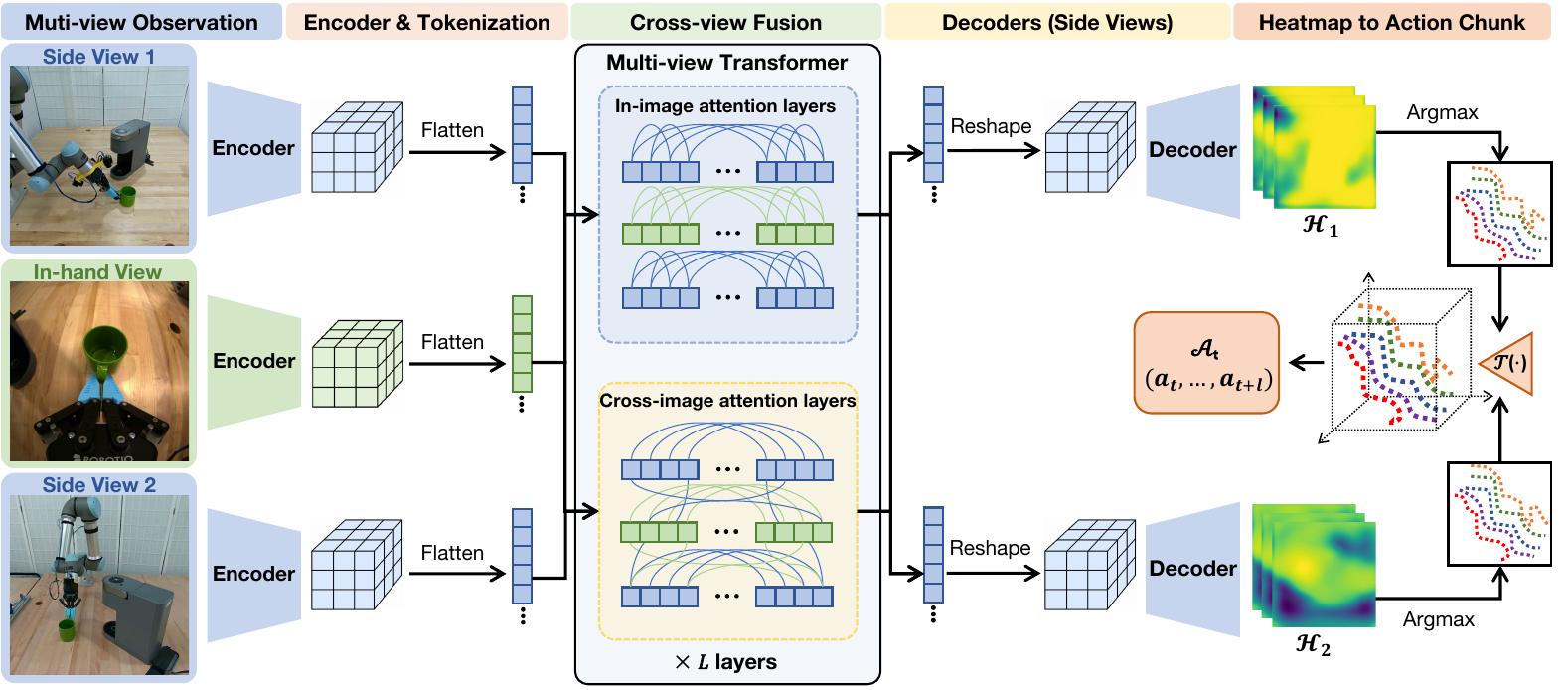}
    \caption{\textbf{Architecture of Action Map Policy.} The left branches take one in-hand image and two side-view images as input. The center features a multi-view transformer that enables communication between the in-hand features and the side-view context features. The right branch consists of two decoders that generate heatmaps of keypoints across the temporal horizon.}
    \label{fig:archi}
    \vspace{-0.2cm}
\end{figure*}
We represent action chunks as spatiotemporal keypoints
and factorize the policy function $\pi$ into two mappings: a heatmap predictor $\psi: \mathcal{O} \rightarrow \mathcal{H}$, 
which maps observations to keypoint distributions, and an action extractor $f: \mathcal{H} \rightarrow \mathcal{A}$, which converts heatmaps into executable actions. The entire policy can be expressed as the composition $\pi = f \circ \psi$.



\begin{wrapfigure}[8]{r}{0.23\textwidth}
    \vspace{-10pt}
    \centering
    \includegraphics[width=0.145\textwidth]{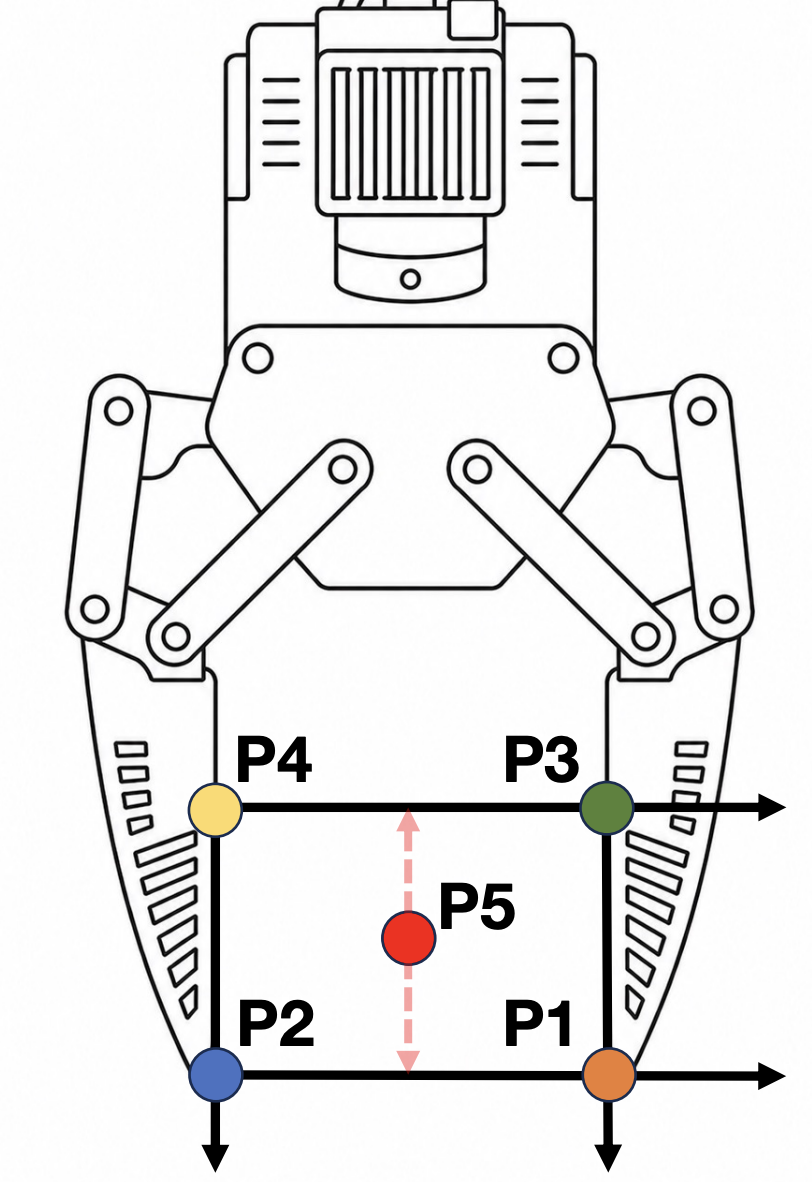}
    \vspace{-5pt}
    \caption{Keypoint-based Action Representation.}
    \label{fig:keypoint_repr}
    \vspace{-5pt}
\end{wrapfigure}

\textbf{Action Chunk Representation as 2D Heatmaps:} Our model is trained entirely in 2D image space to predict actions as 2D pixel keypoints. We first convert 3D pose trajectories into 2D heatmaps to use as labels. Specifically, we map each pose $a = (T,R,w)$ to 3D keypoints $\tilde{a}=(p^1,\ldots, p^m)$ ($m=5$ in our case) using the gripper geometry shown in Figure~\ref{fig:keypoint_repr}. Notably, this choice of keypoints is flexible and extends naturally to other end-effector designs.
Next, we project the 3D keypoints to 2D coordinates in the camera image planes.
Multi-camera projection $\mathcal{P}$ and triangulation $\mathcal{T}$ 
relate 3D points to 2D coordinates via the camera matrices. Under generic camera configurations, $(\mathcal{P}, \mathcal{T})$ establish a bijection between $\mathbb{R}^3$ and $\mathrm{Im}(\mathcal{P}) \subset (\mathbb{R}^2)^n$.
We then \emph{discretize} the projected coordinates into pixel locations and use them to create soft heatmap labels. For keypoint $j$ at timestep $i$ in camera view $k$, the label $\mathbf{h}_{ijk} \in \mathbb{R}^{H \times W}$ is a normalized distribution over the image grid, $\sum_{u,v} \mathbf{h}_{ijk}(u,v) = 1$.

Stacking a single keypoint's heatmaps across the $l$ timesteps along the channel dimension yields its spatiotemporal volume $\mathbf{H}_{jk} \in \mathbb{R}^{l \times H \times W}$. Collecting these over all $m$ keypoints and $n$ cameras transforms the action chunk $\mathcal{A}$ into the set of heatmap volumes $\mathcal{H} = (\mathbf{H}_{jk})_{j=1,k=1}^{m,n}$. Under this representation, the policy function $\pi$ is cast as a dense classification problem over the image plane. The network predicts, for each pixel, the likelihood of being traversed by each action keypoint at each timestep, thereby also unifying observation and action in a shared spatial representation.

\textbf{Predicting Actions as Distributions over Pixels:}
The keypoint-based heatmap formulates actions as a structured set of interaction primitives in image space. We introduce a heatmap predictor $\psi$ to reason over these interactions. Specifically, we introduce what we call the \textit{X-Net} architecture, which features a multi-view image input and multi-view heatmap output design. An example is shown in Figure~\ref{fig:archi}, where the model takes one in-hand image and two side-view images and generates distributions over pixels (normalized heatmaps) for the side views only. The left part of the X-Net uses a U-Net encoder ~\cite{ronneberger2015u}. The resulting dense feature maps are flattened into positionally-encoded tokens and fed into a Multi-View Transformer (MVT)~\cite{goyal2023rvt,hamdi2021mvtn}. The MVT is comprised of in-image attention layers and cross-image attention layers to enable communication across views. The output sequence of tokens is reshaped to recover the spatial dimensions and then fed into the U-Net decoder to generate the heatmap. 
Skip connections are employed between corresponding layers of the encoder and decoder, as is standard for the U-Net architecture. While we illustrate a two camera setup, this architecture is flexible and can be easily adapted to different numbers of cameras. 

The model outputs a heatmap $\mathbf{\hat{h}}_{ijk}$ for each timestep $i$ for each keypoint $j$ in each camera view $k$. The loss is the cross-entropy loss between the predicted heatmap $\mathbf{\hat{h}}_{ijk}$ and the label soft one-hot heatmap $\mathbf{h}_{ijk}$ with a Gaussian distribution over pixels centered at $(u,v)$ obtained by projecting the 3D keypoint
as described above. The loss is averaged across all timesteps, keypoints, and cameras: $\mathcal{L}_{\mathrm{CE}} = - \sum_{i=1}^{l} \sum_{j=1}^{m} \sum_{k=1}^{n} \sum_{u,v}
\mathbf{h}_{ijk}(u,v) \, \log \hat{\mathbf{h}}_{ijk}(u,v).$ 

\textbf{Heatmap-to-Action Conversion:}
We do not extract actions during training; the network is supervised entirely in heatmap space. During inference, we use an action extractor $f: \mathcal{H} \rightarrow \mathcal{A}$ to recover executable actions.
As shown in Figure~\ref{fig:archi}, we first identify the pixel locations with the highest probability via $\arg\max$, yielding $m$ 2D keypoints across $n$ cameras and $l$ time steps, denoted as $\{(u,v)_{ijk}\}$.
We then reconstruct the 3D coordinates of each keypoint via triangulation using the camera matrices, i.e., $\mathcal{T}(\{(u,v)_{ijk}\}_{k=1}^{n}) \mapsto p^j \in \mathbb{R}^3$.

After triangulation, we obtain the trajectory of $m$ keypoints ($m=5$ in our case), from which we recover the per-step pose $(T, R, w)$ as follows.
The translation $T$ is the centroid of the four grasp points, $T = \tfrac{1}{4}(p^1 + p^2 + p^3 + p^4)$.
The orientation $R$ is built from two geometric axes, the antipodal axis $v_{\mathrm{antipodal}} = \tfrac{1}{2}[(p^1 - p^2) + (p^3 - p^4)]$ and the approach axis $v_{\mathrm{approach}} = \tfrac{1}{2}[(p^2 - p^4) + (p^1 - p^3)]$, which we orthonormalize via Gram--Schmidt under the right-hand rule.
The gripper width $w$ is the normalized ratio of $p^5$'s distances to the two fingertip midpoints, $w = d_{+} / (d_{+} + d_{-})$, where $d_{+} = \|p^5 - \tfrac{1}{2}(p^2+p^4)\|$ and $d_{-} = \|p^5 - \tfrac{1}{2}(p^1+p^3)\|$.
\footnote{Points $p^1$--$p^4$ are arranged in a fixed relative configuration for robust axis estimation; consequently, their pairwise distances (e.g., $\|p^3 - p^1\|$) are constant and cannot encode the gripper width.}
By computing the pose at each step, we obtain a sequence of $l$ actions that can directly command the robot. The policy generates new actions by consuming the latest observation at a fixed frequency, yielding a closed-loop visuomotor policy.

\textbf{Implementation Details.}
We report key implementation choices here. Each keypoint is rendered as a Gaussian soft label with $\sigma=2$. The setup uses two side-view cameras and one in-hand camera with $224 \times 224$ images, predicting heatmaps only for the side views. The encoder has four ResBlocks with $16\times$ downsampling, followed by a six-layer MVT module with two intra-image and four cross-image attention layers. The decoder has four upsampling ResBlocks and outputs 60 channels per side view, representing 5 keypoints over 12 time steps; only the first 8 are used during execution. Extensive data augmentation is applied by transforming images and keypoint labels jointly. This is another advantage of our formulation: actions are spatially aligned with observations, enabling equivariant augmentation. Our pseudocode is provided in Appendix~\ref{appendix:pseudocode}. Details on data preprocessing and augmentation
are provided in Appendix~\ref{implementation-details}.

\section{Experiments}
We first measure the theoretical precision that the heatmap action representation can achieve as a function of image resolution, and then evaluate its policy-learning performance in simulation, followed by ablation studies. Finally, we design two types of real-world experiments to test its real-world performance and spatial reasoning ability. Please note that we mainly focus on policy performance \textit{without} using pretrained vision encoder for any method.

\subsection{Precision and Robustness Analysis}
\begin{figure*}[t]
  \centering
  \begin{minipage}[c]{0.51\linewidth}
      \centering
      \includegraphics[width=\linewidth]{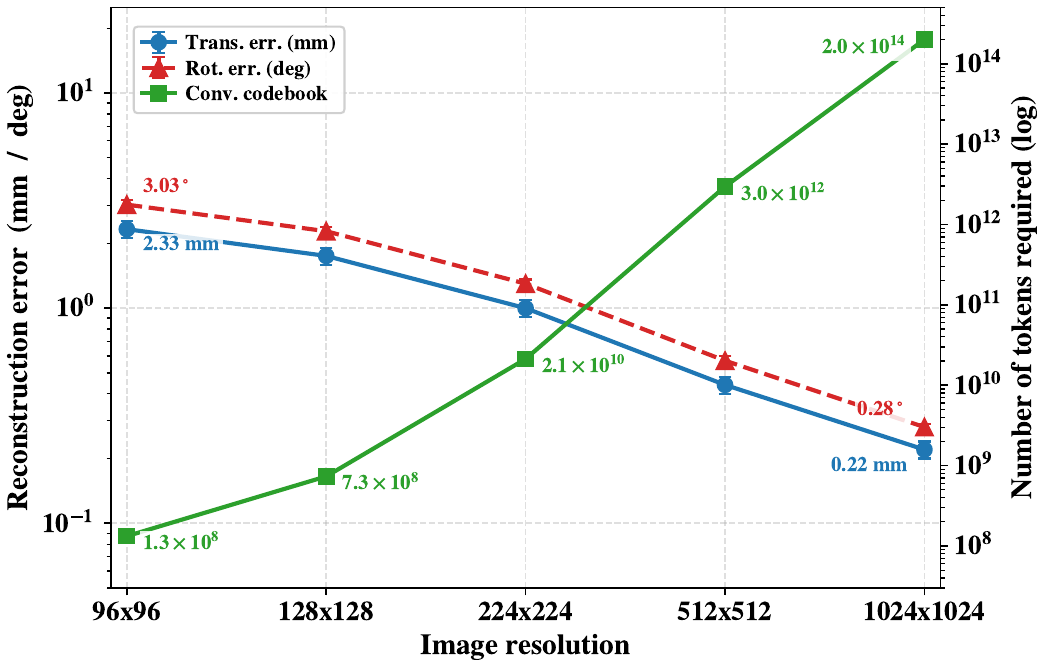}
      \captionsetup{skip=1pt}
      \captionsetup{width=0.9\linewidth}
      \captionof{figure}{Reconstruction error (left) and conventional
  codebook size (right) vs.\ image resolution.
  The codebook grows by six orders of magnitude as resolution increases.}
      \label{fig:reconstruction_precision}
  \end{minipage}
  \hfill
  \begin{minipage}[c]{0.48\linewidth}
      \centering
      \scriptsize
      \setlength{\tabcolsep}{2pt}
      \renewcommand{\arraystretch}{1.80}
      \begin{tabular}{@{}c|c|c|c@{}}
          \toprule
          Resolution & Trans.\ (mm) & Rot.\ ($^\circ$) & \# Tokens \\
          \midrule
          $96 \times 96$       & $2.33\pm 0.21$ & $3.03\pm 0.14$ &
  $(\frac{10}{2.33})^{3}(\frac{360}{3.03})^{3}$ \\
          $128 \times 128$     & $1.75\pm 0.16$ & $2.28\pm 0.11$ &
  $(\frac{10}{1.75})^{3}(\frac{360}{2.28})^{3}$ \\
          $224 \times 224$     & $1.00\pm 0.09$ & $1.30\pm 0.06$ &
  $(\frac{10}{1.00})^{3}(\frac{360}{1.30})^{3}$ \\
          $512 \times 512$     & $0.44\pm 0.04$ & $0.57\pm 0.03$ &
  $(\frac{10}{0.44})^{3}(\frac{360}{0.57})^{3}$ \\
          $1024 \times 1024$   & $0.22\pm 0.02$ & $0.28\pm 0.01$ &
  $(\frac{10}{0.22})^{3}(\frac{360}{0.28})^{3}$ \\
          \bottomrule
      \end{tabular}
      \captionsetup{skip=10pt}
      \captionsetup{width=0.92\linewidth}
      \captionof{table}{Reconstruction accuracy at different image
  resolutions.
      Resolution is shown as $n^2$ for an $n{\times}n$ image; the
  conventional
      discretization uses translation range $10\,\mathrm{mm}$ and
  rotation range
      $360^\circ$.}
      \label{tab:reconstruction_precision}
  \end{minipage}
  \end{figure*}

To measure the discretization effects of keypoint heatmaps,
we first implement projection and triangulation $(\mathcal{P}, \mathcal{T})$ in continuous space, which can be viewed as a forward transform $\mathcal{P}$ and its inverse $\mathcal{T}=\mathcal{P}^{-1}$. The composition $\mathcal{T}\cdot\mathcal{P}$ is the identity mapping, even under imperfect camera calibration\footnote{Robustness analysis: While real-world calibration is imperfect and introduces small offsets, the transformation remains effectively invertible, and the policy model naturally adapts to such projected data during training.}. We then discretize the continuous image coordinates into pixels, yielding $\mathcal{T}\cdot\hat{\mathcal{P}}$, where $\hat{\mathcal{P}}$ denotes a discretized projection operator determined by the image resolution. Next, we map the ground-truth action $a_{\mathrm{gt}} = (T, R, w)$ to a set of pixel coordinates, reconstruct the corresponding 3D keypoints, and compute the reconstructed action $a_{\mathrm{rec}}$.
We evaluate reconstruction precision on our real-world data in Table~\ref{tab:reconstruction_precision}, measuring both translation and rotation errors. We compare it with conventional discretization methods that bin each action dimension and report the number of tokens required to achieve comparable accuracy.

As shown in Table~\ref{tab:reconstruction_precision}, with a $224 \times 224$ image resolution, the heatmap representation achieves approximately $1\,\mathrm{mm}$ translation precision and $1.3^\circ$ rotation precision. In contrast, conventional discretization would require around $10^{10}$ codebook tokens to achieve comparable precision, resulting in a prohibitively large number of classes. Moreover, the precision improves approximately linearly as the image resolution increases, achieving a translation error of $0.22\,\mathrm{mm}$ and a rotation error of $0.28^\circ$ at a resolution of $1024 \times 1024$. In contrast, the number of required tokens for conventional discretization grows polynomially as $n^6$, leading to a rapidly increasing codebook size. This precision provides solid theoretical support for keypoint-heatmap-based policy learning.



\subsection{Simulated Experiments}
We evaluate $\ours$ on six representative tasks from MimicGen~\cite{mandlekar2023mimicgen}. These tasks cover both rigid and articulated objects, with medium to high precision requirements.

\begin{figure}[!b]
    \centering
    \setlength{\tabcolsep}{2pt}
    \renewcommand{\arraystretch}{0}
    \begin{tabular}{cccccc}
        \includegraphics[width=0.15\textwidth]{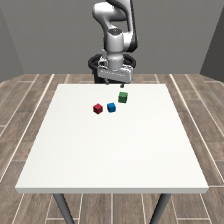} &
        \includegraphics[width=0.15\textwidth]{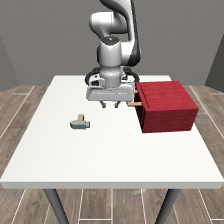} &
        \includegraphics[width=0.15\textwidth]{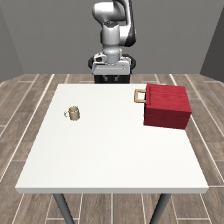} &
        \includegraphics[width=0.15\textwidth]{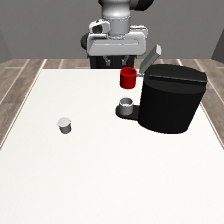} &
        \includegraphics[width=0.15\textwidth]{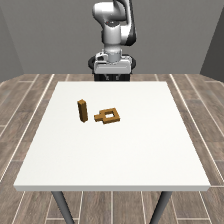} &
        \includegraphics[width=0.15\textwidth]{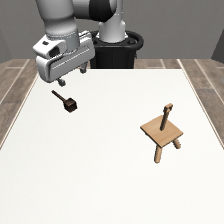} \\
    \end{tabular}
    \captionsetup{width=0.95\linewidth}
    \caption{Representative tasks from MimicGen. From left to right: {stack-three-block}, {hammer-cleanup-d1}, {mug-cleanup-d1}, {coffee-d2}, {square-d2}, and {threading-d2}.}
    \label{fig:task}
    \vspace{-0em}
\end{figure}

\textbf{Task Description.}
As shown in Figure~\ref{fig:task}, stack-three-d1 requires the robot to stack three blocks; hammer-cleanup-d1 and mug-cleanup-d1 require the robot to open a drawer, put the target inside, and close the drawer; coffee-d2 is to place a coffee pod into the machine, while square-d2 and threading-d2 feature high-precision insertion tasks. The suffixes d0, d1, and d2 indicate the degree of spatial variation in the test scenarios, ranging from light to substantial variation. We use the largest initial-state distribution provided by MimicGen~\cite{mandlekar2023mimicgen}, e.g., d2 for coffee and d1 for stack. We use two side views and one in-hand view for each task. The camera views are provided in the Appendix~\ref{simulation_details}.



\begin{table*}[!b]
  \setlength\tabcolsep{6pt}
  \begin{center}
  \scriptsize
  \begin{tabular}{@{}lcccccccc@{}}
  \toprule
  & stack-three-d1 & hammer-cleanup-d1 & mug-cleanup-d1 & coffee-d2 & square-d2 & threading-d2 \\
  \cmidrule(lr){2-2} \cmidrule(lr){3-3} \cmidrule(lr){4-4} \cmidrule(lr){5-5} \cmidrule(lr){6-6} \cmidrule(lr){7-7}
  Method & 100 Demo & 100 Demo & 100 Demo & 100 Demo & 100 Demo & 100 Demo \\
  \midrule
  Diffusion Policy~\cite{chi2025diffusion} & 38 & 58 & \textbf{60} & 62 &20 & 26 \\
  ACT~\cite{zhao2023learning} & 14 & 60 & 44 & 42 & 6 & 20   \\
  OAT~\cite{liu2026oat} & 34 & 32 & 22 & 30 & 18 & 16 \\
  Motion Track~\cite{ren2025motion} & 8 & 40 & 24 & 36 & 14 & 12 \\
  $\ours$ & \textbf{90} & \textbf{88} & 52 & \textbf{78} & \textbf{50} & \textbf{30} & \\

  \bottomrule
  \end{tabular}
  \end{center}
  \vspace{-0.5em}\
  \captionsetup{width=0.95\linewidth}
  \caption{{Performance Comparisons on Mimicgen Benchmark.} Success rate (mean\%) over 50 unseen tests with 100 demonstration episodes used for training.}
  \label{table:sim-results}
\end{table*}

\textbf{Baseline.}
We compare AMP against four strong baselines that cover different policy-learning frameworks: (1) DiffPo~\cite{chi2025diffusion}, which uses an iterative denoising approach to predict actions conditioned on extracted observation embeddings; (2) ACT~\cite{zhao2023learning}, which directly maps observations to actions and is trained with an MSE loss, using a pretrained CVAE to represent the action-style vector; (3) OAT~\cite{liu2026oat}, a token-based method that first learns a codebook with an autoencoder on demonstration actions and then predicts action tokens; and (4) Motion Track~\cite{ren2025motion}, a diffusion-based policy variant that takes a single view as input, denoises pixel keypoint coordinates independently, and triangulates the denoised coordinates from two views. We evaluate each method 20 times during training and report the best performance over 50 tests across checkpoints.

\textbf{Results.} We report our result in Table~\ref{table:sim-results}. Several conclusions can be drawn from it. (1) Action Map Policy outperforms different learning methods on five of the six tasks, with an average gain of $20.7\%$ over the second-best baseline, DiffPo ($44.0\%$). The improvement on individual tasks ranges from $4\%$ on threading-d2 to $52\%$ on stack-three-d1, demonstrating its advantage as a general policy-learning framework. (2) Compared with the regression-based method ACT and the token-based method OAT, AMP achieves better performance across all tasks without requiring any action pretraining. (3) AMP largely outperforms the keypoint-based diffusion method Motion Track. We find that Motion Track can produce inconsistent trajectories due to its independent inference strategy, whereas AMP maintains consistent predictions through its cross-view design.

\begin{wraptable}[5]{r}{0.53\linewidth}
\vspace{-0.7em}
\centering
\scriptsize
\setlength{\tabcolsep}{2pt}
\renewcommand{\arraystretch}{1}
\resizebox{\linewidth}{!}{
\begin{tabular}{l c c c}
\toprule
Method & stack-three-d1 & hammer-cleanup-d1 & coffee-d2 \\
\midrule
Action Map Policy & 90 & 88 & 78 \\
w/o in-hand camera & 88 ($\downarrow$2) & 74 ($\downarrow$14) & 64 ($\downarrow$14) \\
w/o soft labels ($\sigma=0$) & 82 ($\downarrow$8) & 78 ($\downarrow$10) &  70 ($\downarrow$8)\\
\bottomrule
\end{tabular}
}
\vspace{-0.5em}
\caption{Ablation study results.}
\label{tab:ablation}
\end{wraptable}
\textbf{Ablation Study.}
We conduct ablation studies by removing the in-hand camera view and by using one-hot keypoint labels instead of soft labels. We report the results in Table~\ref{tab:ablation}. Removing the in-hand camera leads to a performance drop on tasks with heavier occlusion, suggesting that the in-hand view provides complementary visual information. In addition, using soft labels improves performance by 10\% compared to one-hot labels. Additional ablation study can be found in Appendix~\ref{equiv_data_aug}.


\subsection{Real-World Experiments}
\begin{figure*}[h]
\centering
\begin{subfigure}[t]{0.307\linewidth}
    \centering
    \includegraphics[width=\linewidth]{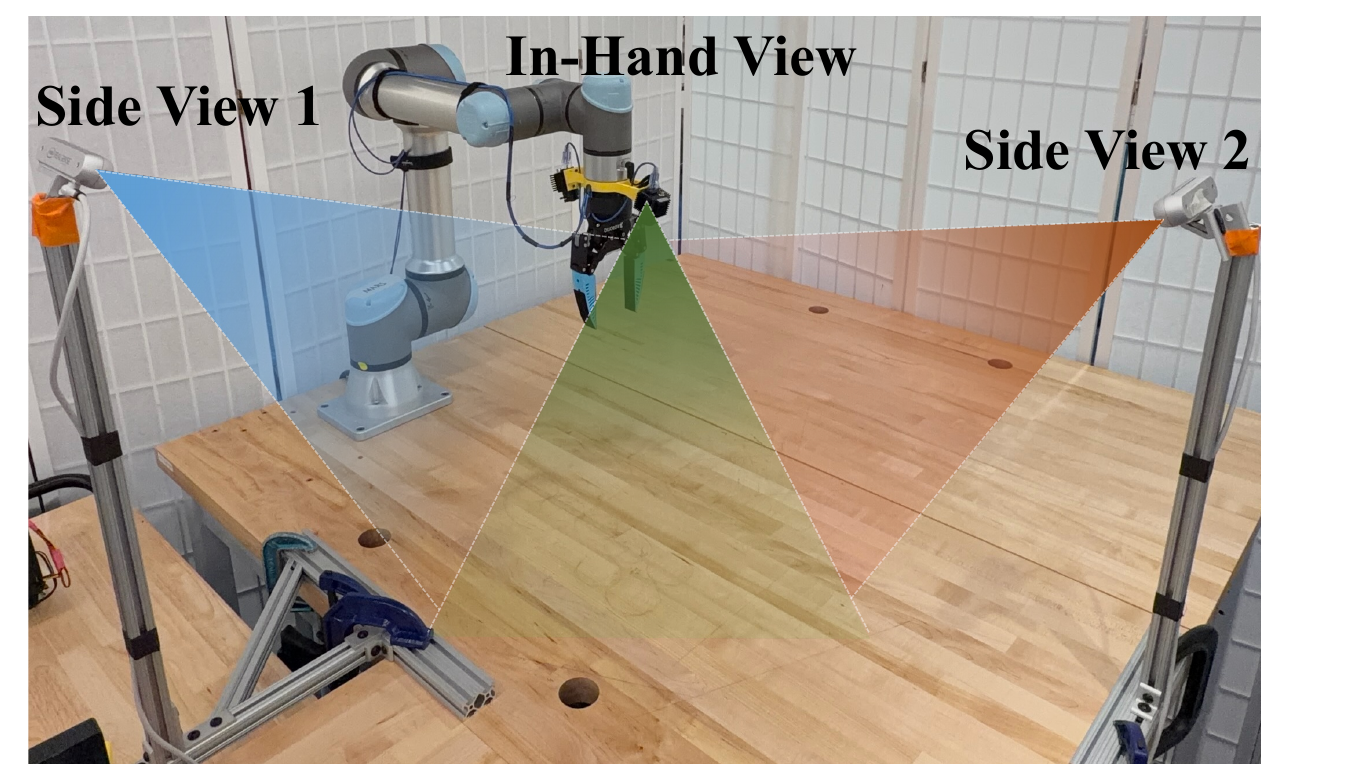}
    \caption{workspace setup}
    \label{fig:real_setup_a}
\end{subfigure}
\hspace{-0.02\linewidth}
\begin{subfigure}[t]{0.17\linewidth}
    \centering
    \includegraphics[width=\linewidth]{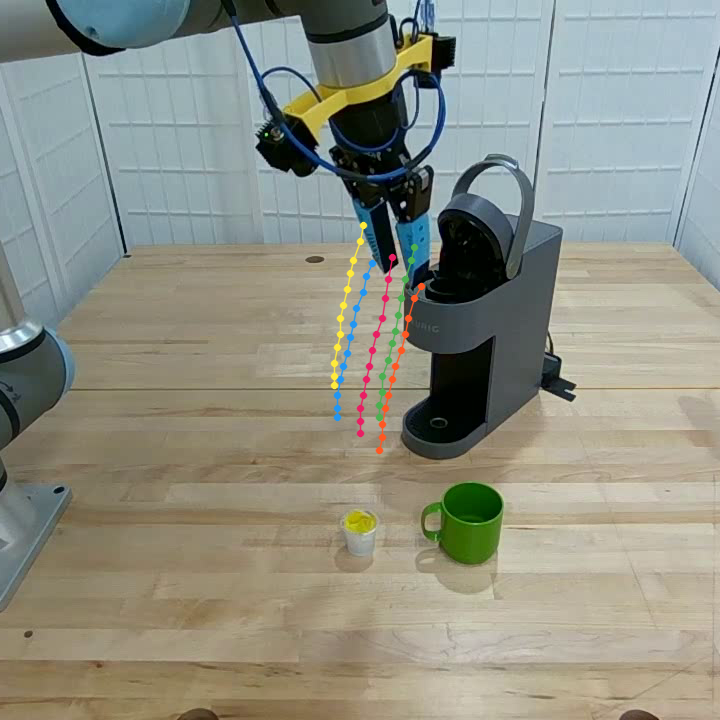}
    \caption{sideview one}
    \label{fig:real_setup_b}
\end{subfigure}
\hfill
\begin{subfigure}[t]{0.17\linewidth}
    \centering
    \includegraphics[width=\linewidth]{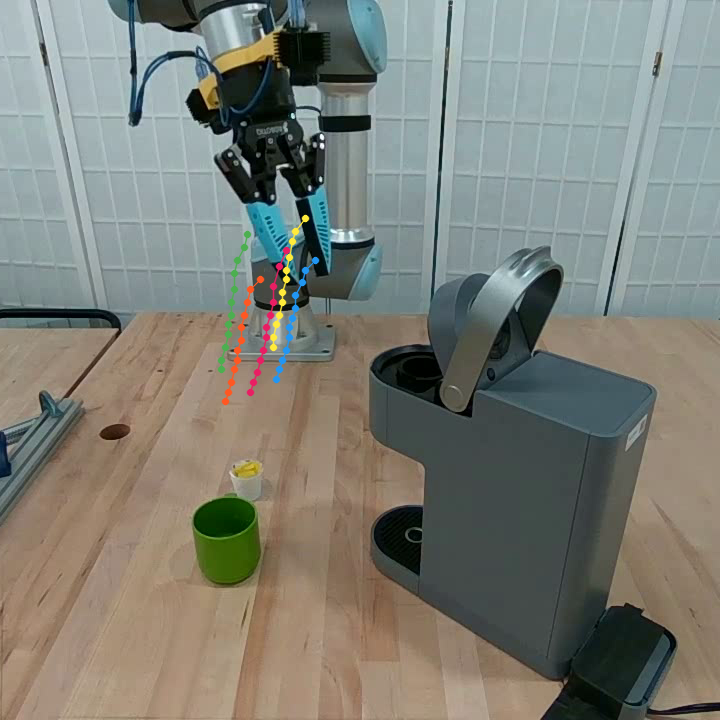}
    \caption{sideview two}
    \label{fig:real_setup_c}
\end{subfigure}
\hfill
\begin{subfigure}[t]{0.17\linewidth}
    \centering
    \includegraphics[width=\linewidth]{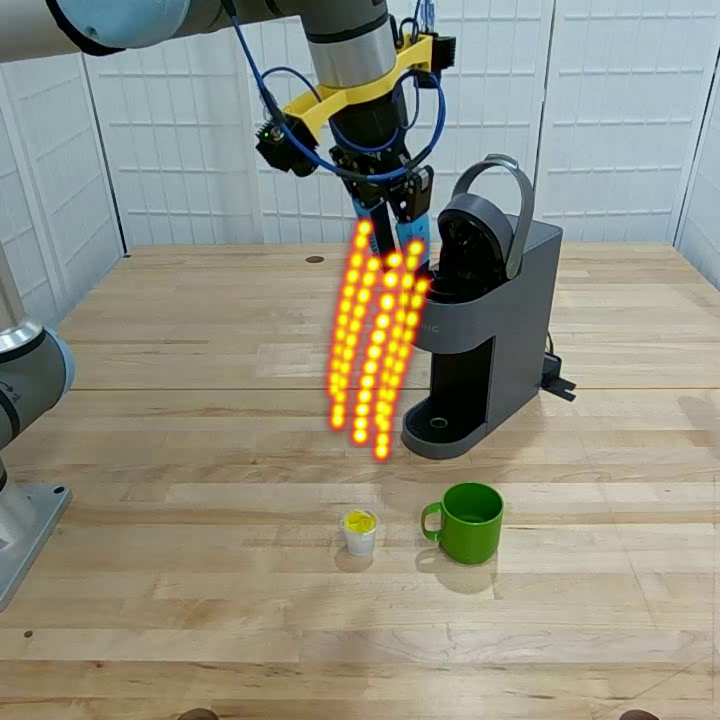}
    \caption{keypoint prediction}
    \label{fig:real_setup_d}
\end{subfigure}
\hfill
\begin{subfigure}[t]{0.17\linewidth}
    \centering
    \includegraphics[width=\linewidth]{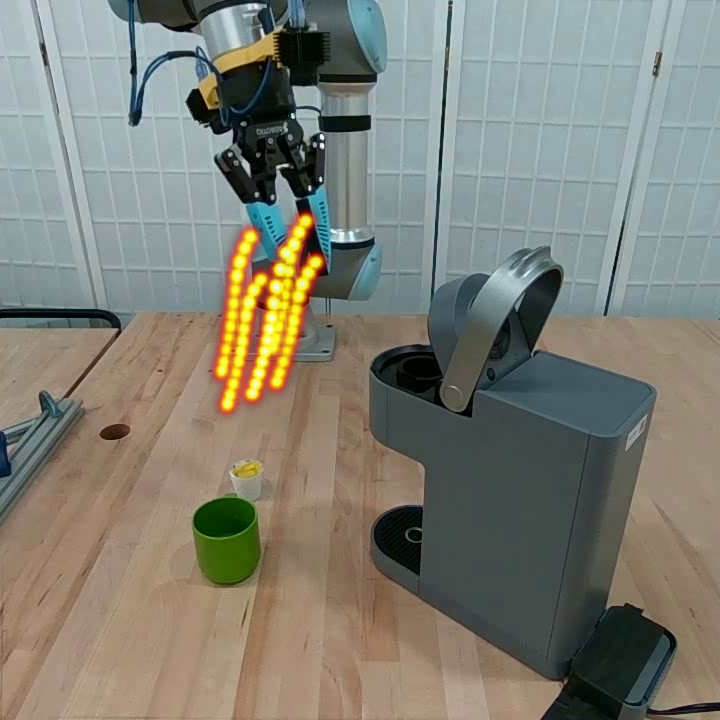}
    \caption{keypoint prediction}
    \label{fig:real_setup_e}
\end{subfigure}

\caption{Real-world system setup, ground-truth keypoint projection visualization, and action map prediction.}
\label{fig:real_world_setup}
\end{figure*}

We conduct five real-world experiments to further evaluate our method as a closed-loop visuomotor policy-learning approach and test its spatial reasoning ability. The system setup is shown in Figure~\ref{fig:real_setup_a}. We use a UR5 arm equipped with dual in-hand cameras, of which only one is used, and two RealSense D455 cameras as side views pointing toward the workspace. We calibrate the side-view cameras, collect demonstrations using Gello~\cite{wu2024gello}, and project the keypoint trajectories onto the images, as shown in Figure~\ref{fig:real_setup_b} and \ref{fig:real_setup_c}.

\begin{figure*}[t]
  \centering
  \setlength{\abovecaptionskip}{4pt}
  \setlength{\belowcaptionskip}{0pt}

  \includegraphics[width=0.33\linewidth]{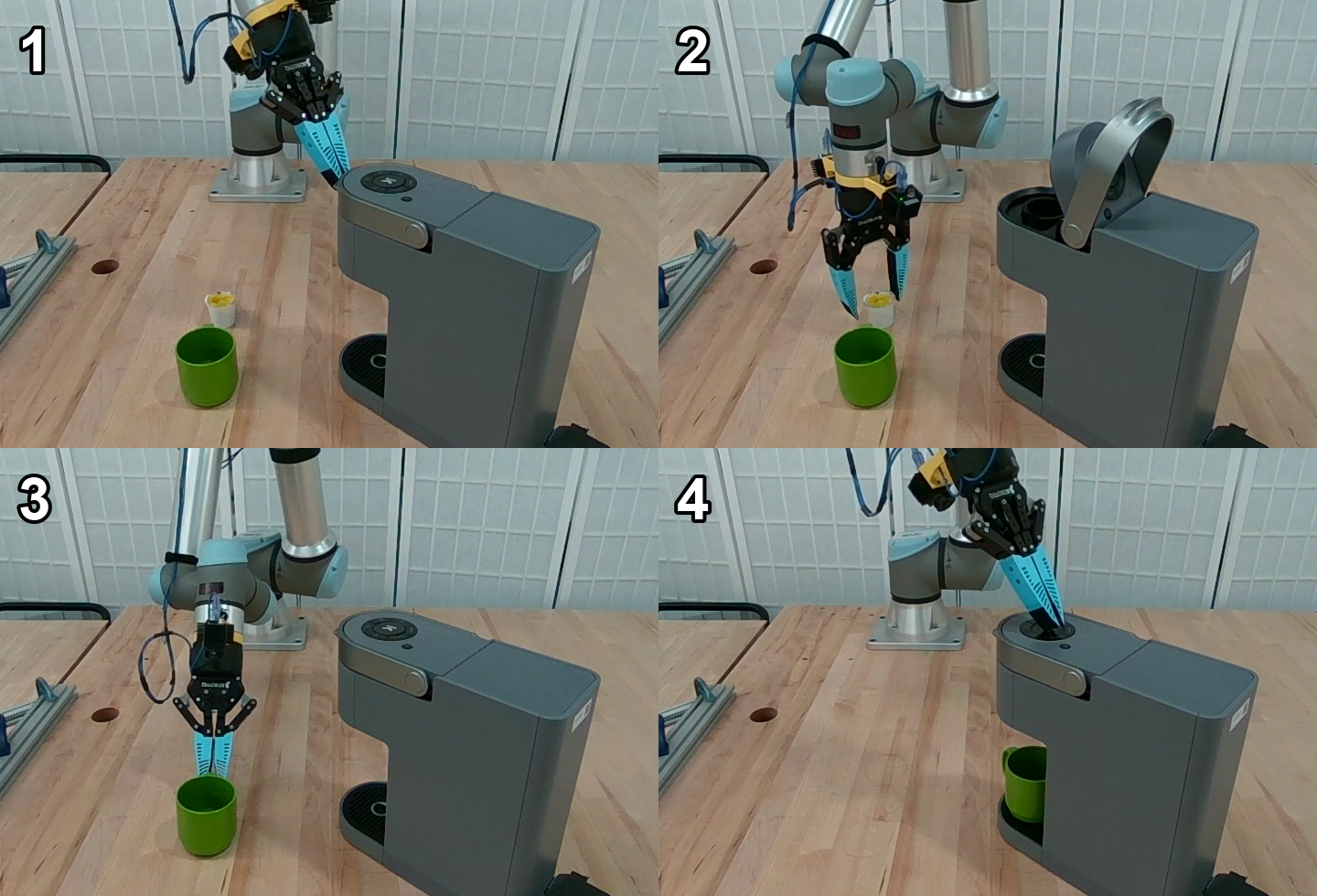}%
  \hfill
  \includegraphics[width=0.33\linewidth]{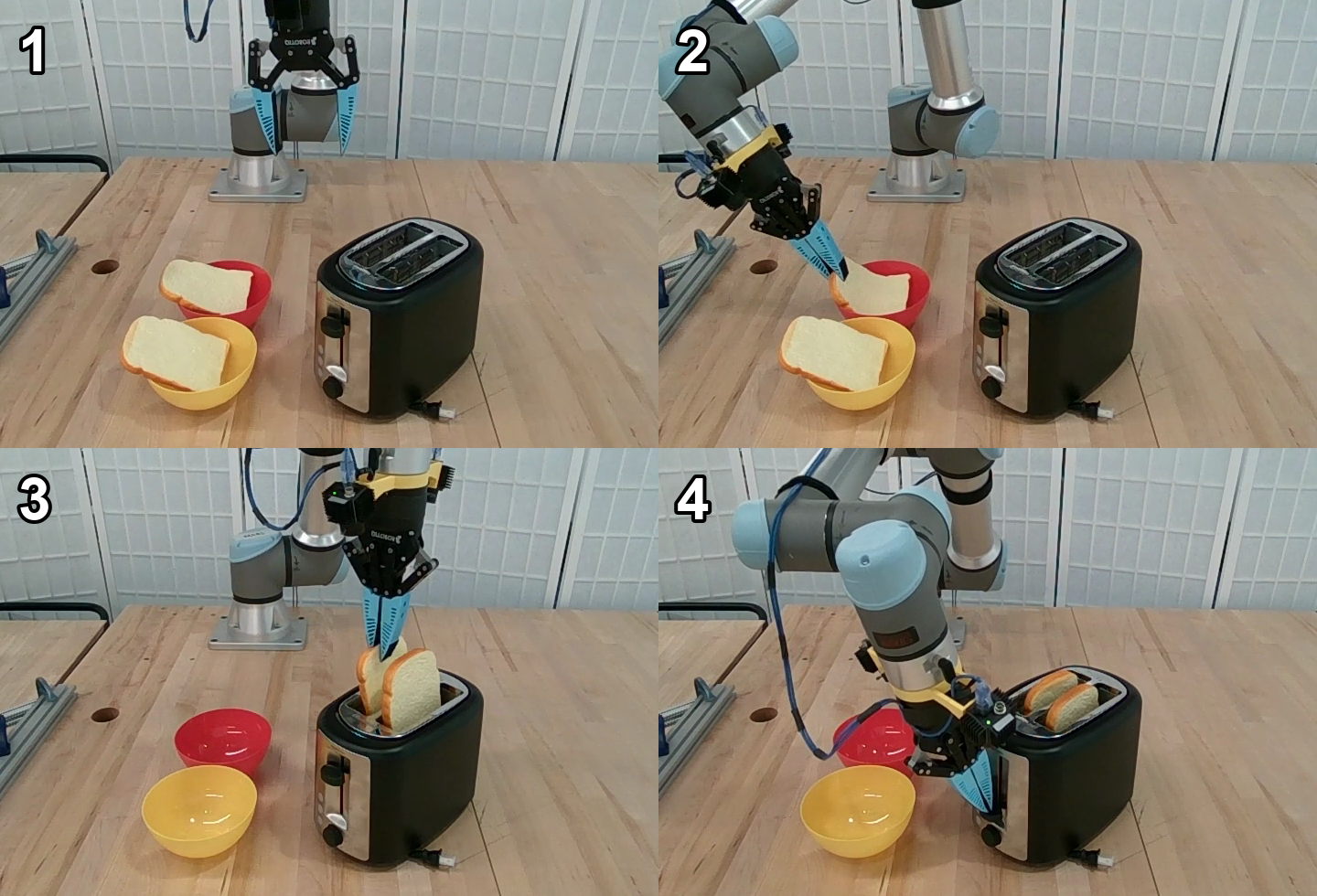}%
  \hfill
  \includegraphics[width=0.33\linewidth]{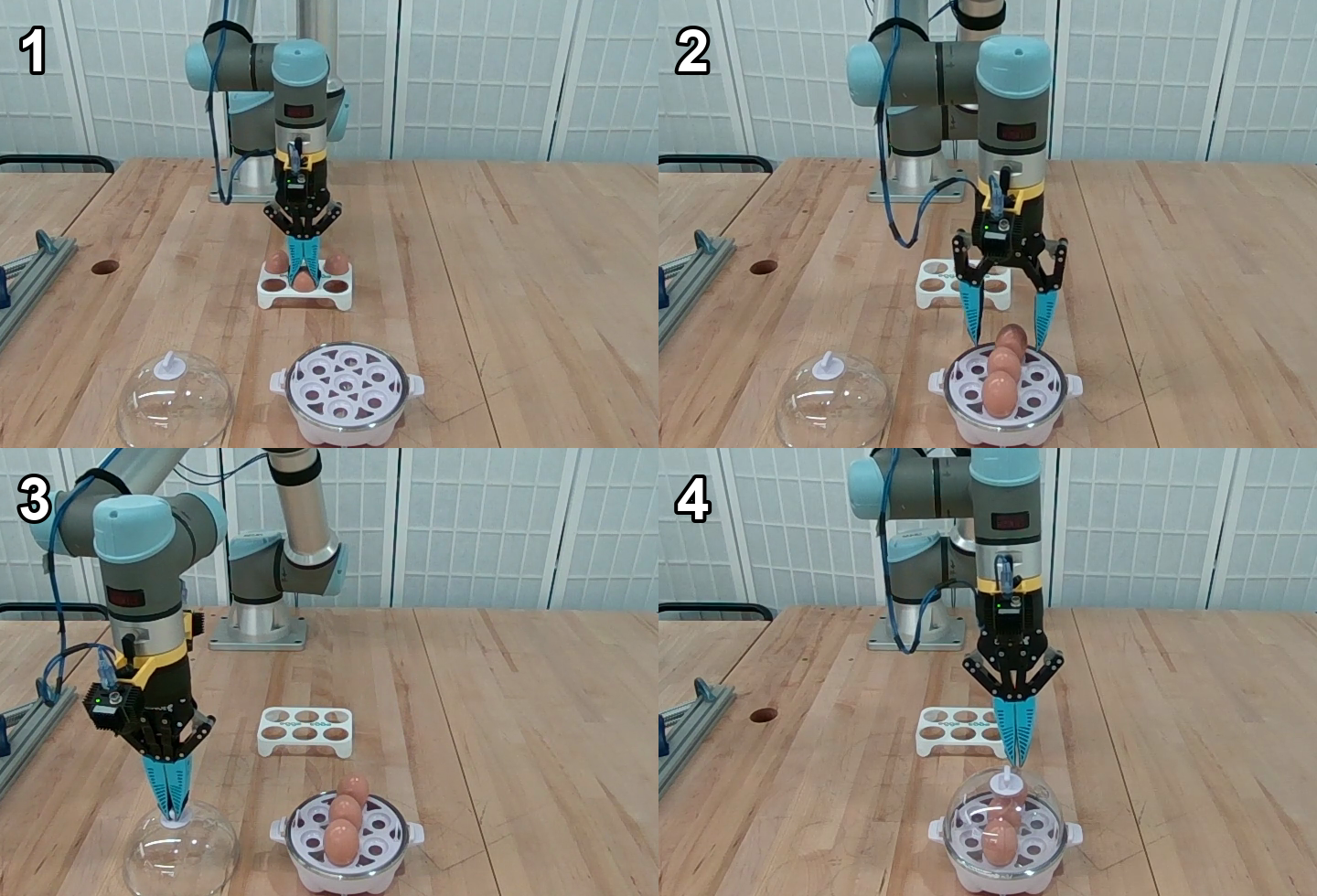}

  \caption{Real-world tasks: make-coffee, toast-bread, and
  steam-egg.}
  \label{fig:realworld_tasks_1}
  \end{figure*}

\textbf{Real-World Evaluation of Visuomotor Policy Learning.} We design three challenging long-horizon real-world tasks and compare our method with two strong baselines, DiffPo and ACT. As shown in Figure~\ref{fig:realworld_tasks_1}, (1) making-coffee requires the robot to open the coffee machine lid, insert a coffee capsule, place the mug, and press the button. (2) toast-bread requires the robot to place each bread slice into the toaster slot and press the lever down. (3) steam-egg requires the robot to iteratively place three eggs into the slots and place the lid onto the egg steamer. For each task, we collect 70 demonstrations and evaluate each method over 20 runs with varied initial object placements. These tasks are long-horizon and require precise manipulation, and we report the task completion rates in Table~\ref{tab:realworld_long_horizon}.  We find that (1) AMP outperforms DiffPo~\cite{chi2025diffusion} and ACT~\cite{zhao2023learning} by a margin of \textbf{50\%--70\%}. We observe that the baselines degrade significantly under spatial variations (Figure~\ref{fig:realworld_long_horizon}) and precision-demanding settings, whereas our method maintains strong spatial generalization; (2) Compared with iterative methods, AMP generates actions with a single forward pass ($13.80\,\mathrm{ms}$), making it substantially faster than DiffPo ($93.53\,\mathrm{ms}$), which uses 16 DDIM steps. Additional details are provided in Appendix~\ref{appendix:heatmap_vis}.

\textbf{Real-World Evaluation of Spatial Reasoning.} We tailor two tasks, including grasp-target-cup (transparent), and stack-target-block, to evaluate the spatial reasoning ability of our method and baselines. Specifically, we present four identical objects, and a human uses a laser pointer to indicate the target object to manipulate. As shown in Figure~\ref{fig:realworld_tailored_tasks}, these tasks feature small laser-point cues and multi-modal object distributions caused by identical objects. We collect 40 demonstrations for each task with small variations and evaluate each model over 20 test runs. We report the results in Table~\ref{tab:realworld_tailored_tasks}. Our method achieves a success rate of 100\%, while DiffPo only achieves an average of 20\%. We observe two main failure cases for the baselines: (1) the policy ignores the laser point and grasps a non-target object; and (2) the policy approaches the target object but fails to manipulate it due to incorrect action prediction or collapses toward the middle location between two identical objects. We hypothesize that the root cause is that DiffPo denoises actions \emph{conditioned} on observation features, which may make it difficult to directly respond to fine-grained geometric cues. ACT directly maps observations to actions and shows better spatial reasoning ability than Diffusion Policy. However, its regression-based formulation may collapse multi-modal behaviors, especially when identical objects are present. In contrast, our method captures fine-grained observation signals and responds correctly, while expressing complex action distributions through a classification objective.

\begin{table*}[t]
\centering

\begin{minipage}[t]{0.55\textwidth}
\vspace{0pt}
\centering

\scriptsize
\setlength{\tabcolsep}{3.5pt}
\renewcommand{\arraystretch}{1.15}

\begin{tabular}{lcccc}
\toprule
Method 
& Speed
& Coffee
& Toast
& Egg \\
\midrule
DiffPo (DDIM) & 93.53 ms & 25\% (5/20) & 40\% (8/20) & 25\% (5/20) \\
ACT           & 7.16 ms & 15\% (3/20) & 35\% (7/20) & 15\% (3/20) \\
$\ours$       & 13.80 ms & \textbf{80\%} (16/20) & \textbf{90\%} (18/20) & \textbf{85\%} (17/20) \\
\bottomrule
\end{tabular}

\caption{Real-world performance on three long-horizon tasks, evaluated by completion rate and inference speed.}
\label{tab:realworld_long_horizon}

\end{minipage}
\hfill
\begin{minipage}[t]{0.42\textwidth}
\vspace{0pt}
\centering

\includegraphics[width=\linewidth]
{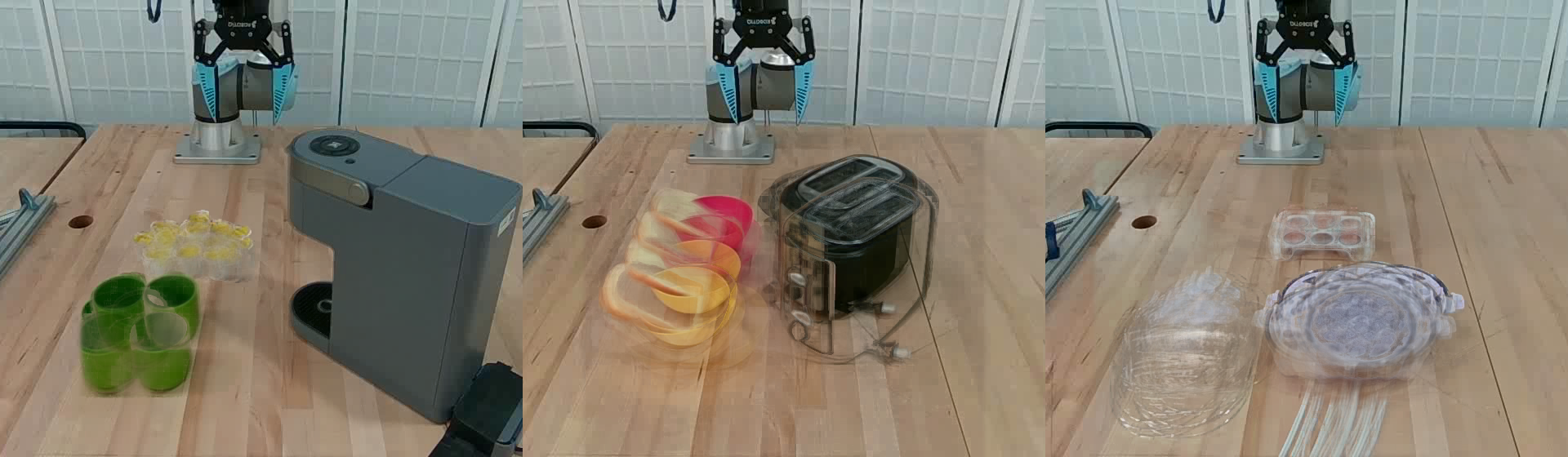}

\vspace{-1pt}

\captionof{figure}{Variation in initial object placements during testing}
\label{fig:realworld_long_horizon}

\end{minipage}

\end{table*}
\begin{figure*}[h]
\centering
\captionsetup{skip=2pt}
\begin{minipage}[t]{0.55\linewidth}
\centering
\vspace{8pt}
\scriptsize
\setlength{\tabcolsep}{5pt}
\renewcommand{\arraystretch}{1.15}
{
\begin{tabular}{l c c}
\toprule
Method & stack-target-cup & stack-target-block \\
\midrule
DiffPo (DDIM) & 25\% (5/20) & 15\% (3/20) \\
ACT & 75\% (15/20) & 65\% (13/20)\\
$\ours$ & \textbf{100\%} (20/20) & \textbf{100\%} (20/20) \\
\bottomrule
\end{tabular}
}
\captionsetup{type=table, skip=4pt}
\caption{Real-world spatial reasoning results.}
\label{tab:realworld_tailored_tasks}
\end{minipage}
\hfill
\begin{minipage}[t]{0.44\linewidth}
\centering
\vspace{0pt}
\begin{subfigure}[t]{0.46\linewidth}
    \centering
    \includegraphics[width=\linewidth]{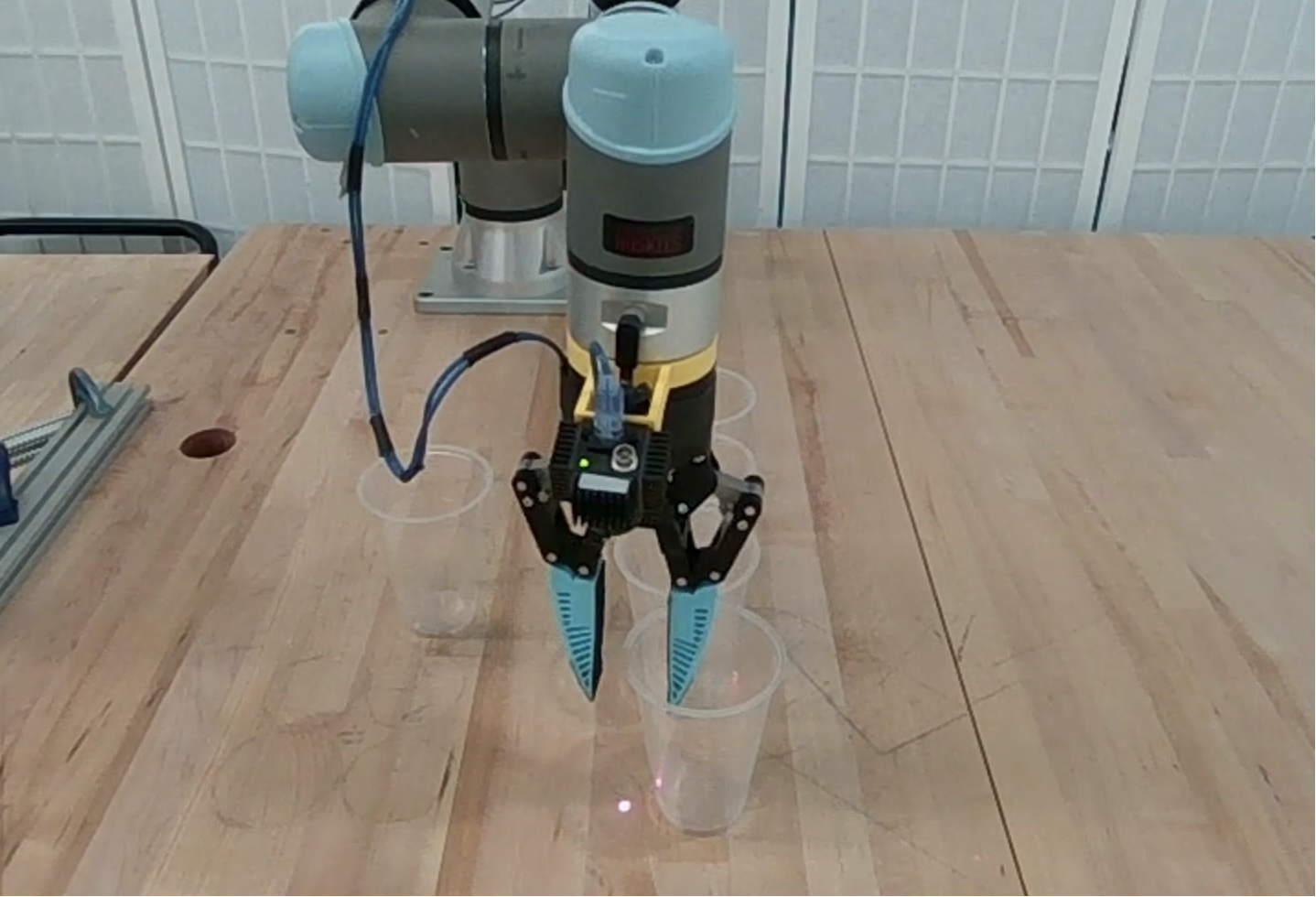}
\end{subfigure}
\hfill
\begin{subfigure}[t]{0.46\linewidth}
    \centering
    \includegraphics[width=\linewidth]{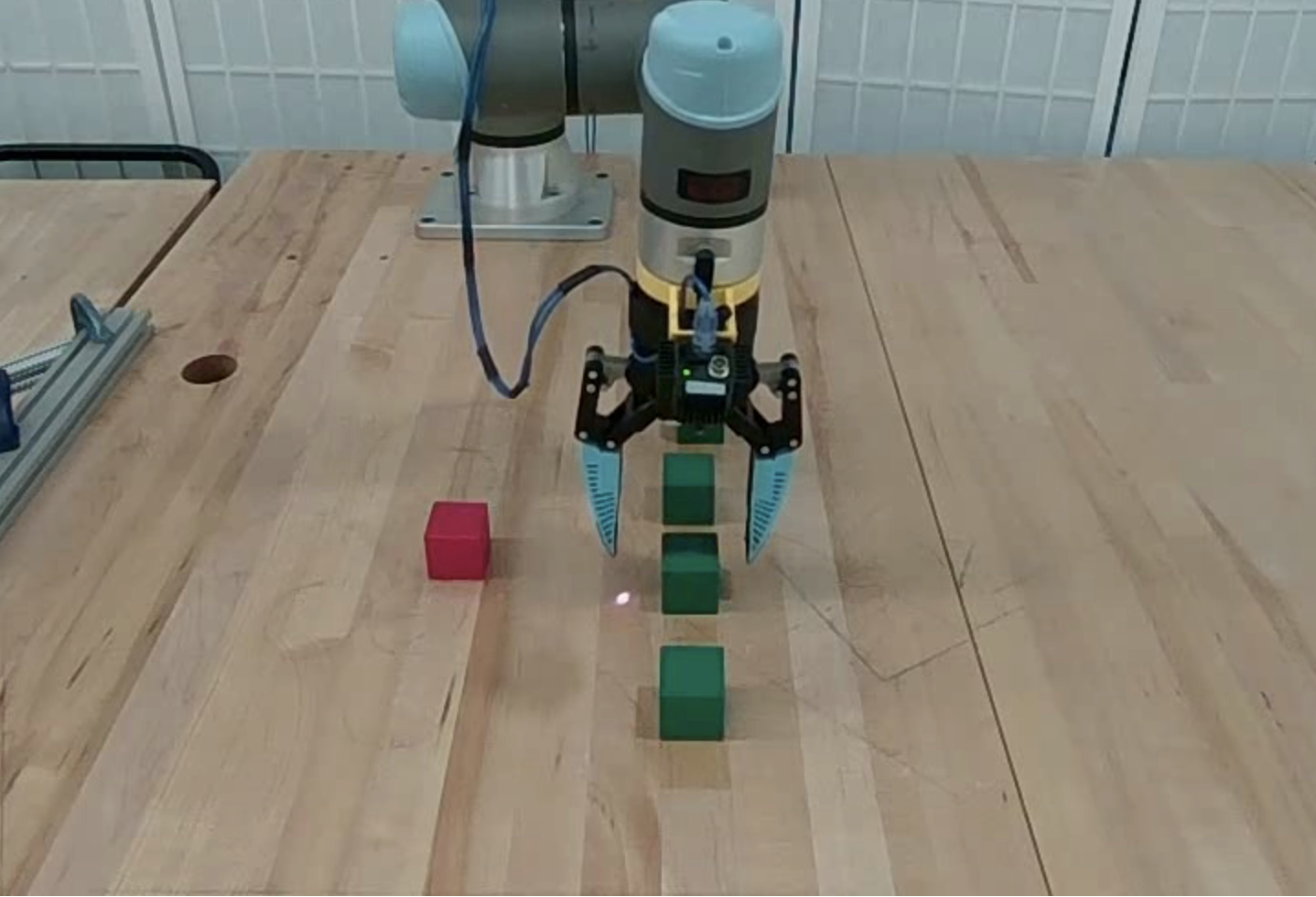}
\end{subfigure}
\captionsetup{type=figure, skip=2pt}
\captionsetup{width=1.04\linewidth}
\caption{Real-world spatial reasoning tasks: stack-target-transparent cup and stack-target-block.}
\label{fig:realworld_tailored_tasks}
\end{minipage}
\end{figure*}



\section{Conclusion and Limitation}
In this work, we propose a novel policy-learning framework that casts closed-loop action learning as classification using action keypoint heatmaps.
Experiments show that AMP significantly outperforms several baselines, while enabling fast single-pass inference and strong responsiveness to fine-grained geometric cues. We also analyze its theoretical precision and show that it can achieve millimeter-level accuracy. This supports our design principle of aligning action precision with observation precision, since such accuracy would not be achievable if the sensors could not capture the corresponding fine-grained information. There are several future directions we plan to explore. We observe that the generated actions are generally smooth, but still exhibit minor jittering. This could potentially be improved by adding a geometric consistency loss to stabilize the predictions. 
Moreover, the workspace is limited by the calibrated camera coverage, which prevents actions outside the observed scope. This could be alleviated with larger-FOV cameras or dual in-hand cameras for dense feature prediction.
While we focus on a parallel-jaw gripper, the method extends naturally to dexterous or bimanual setups by adapting the keypoint design.
In addition, since AMP generates an explicit action distribution, it provides a promising opportunity to learn Q functions for reinforcement learning.
Finally, the classification-based training objective aligns naturally with vision-language models, offering a unified interface for jointly formulating vision, language, and action. While our experiments train only from scratch, this also opens an interesting direction for exploring how pretrained models could further strengthen the framework.

\clearpage
\acknowledgments{This work was supported in part by the National Science Foundation under grants NSF 2442658 and NSF 2314182.
}


\bibliography{example}  

\clearpage
\section{Appendix}

\subsection{Simulation Details}
\label{simulation_details}
\begin{figure}[htbp]
    \centering
    \setlength{\tabcolsep}{2pt}
    \renewcommand{\arraystretch}{0}
    \begin{tabular}{cccccc}
        \includegraphics[width=0.155\textwidth]{figures/mimig_gen_task_figure/stack_three_d1/agentview.png} &
        \includegraphics[width=0.155\textwidth]{figures/mimig_gen_task_figure/hammer_cleanup_d1/agentview.png} &
        \includegraphics[width=0.155\textwidth]{figures/mimig_gen_task_figure/mug_cleanup_d1/agentview.png} &
        \includegraphics[width=0.155\textwidth]{figures/mimig_gen_task_figure/coffee_d2/agentview.png} &
        \includegraphics[width=0.155\textwidth]{figures/mimig_gen_task_figure/square_d2/agentview.png} &
        \includegraphics[width=0.155\textwidth]{figures/mimig_gen_task_figure/threading_d2/agentview.png} \\
        \includegraphics[width=0.155\textwidth]{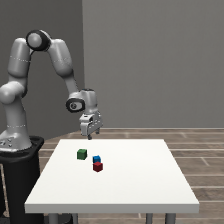} &
        \includegraphics[width=0.155\textwidth]{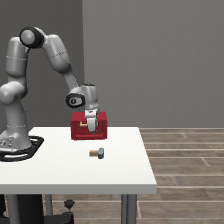} &
        \includegraphics[width=0.155\textwidth]{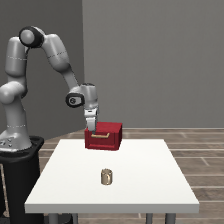} &
        \includegraphics[width=0.155\textwidth]{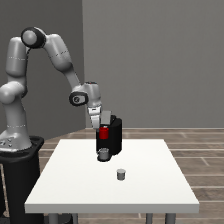} &
        \includegraphics[width=0.155\textwidth]{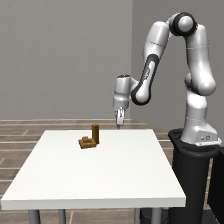} &
        \includegraphics[width=0.155\textwidth]{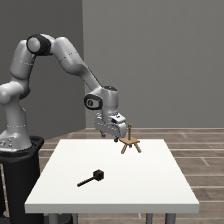} \\
        \includegraphics[width=0.155\textwidth]{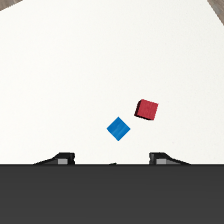} &
        \includegraphics[width=0.155\textwidth]{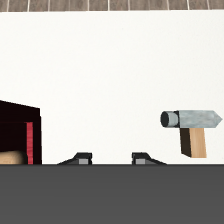} &
        \includegraphics[width=0.155\textwidth]{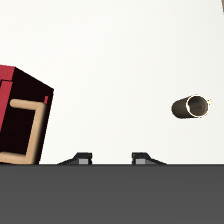} &
        \includegraphics[width=0.155\textwidth]{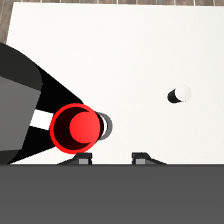} &
        \includegraphics[width=0.155\textwidth]{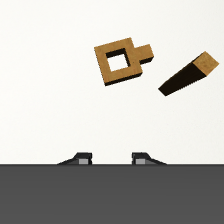} &
        \includegraphics[width=0.155\textwidth]{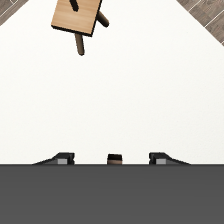} \\
    \end{tabular}
    \caption{
Simulation camera setup for the six MimicGen tasks. The first row shows side-view camera 1, the second row shows side-view camera 2, and the last row shows the in-hand view. All images are rendered at a resolution of $224 \times 224$.
}
    \label{fig:grid_vis}
\end{figure}

\subsection{Implementation Details.}
\label{implementation-details}
To keep the action keypoints within the image space during training, we truncate each action chunk at the first out-of-scope keypoint and replace the remaining keypoint configurations with the last valid in-scope configuration. We also apply extensive data augmentation to AMP. Since the actions are represented in the same image space, each image can be jointly rotated and translated together with its corresponding action heatmap labels. Specifically, for each view, including side view 1, side view 2, and the in-hand view, we apply a random rotation and translation with probability 0.5, transforming the labels accordingly. Rotations are sampled uniformly from $[-\pi/6, \pi/6]$, and translations are sampled uniformly from $[-H/6, H/6]$, where $H$ is the image size.
This exposes the model to multiple transformed variants of similar training samples, improving robustness and reducing overfitting. We discuss the benefits of equivariant data augmentation in more detail in the next subsection. Our transformer consists of 6 attention layers with 8 heads each and a hidden dimension of 768, yielding approximately 41.2M parameters in total. We train the model for 200 epochs on a single NVIDIA RTX 5090 GPU, using a learning rate of $1\mathrm{e}{-4}$ with a cosine annealing schedule.

\begin{wraptable}{r}{0.5\textwidth}
\centering
\scriptsize
\setlength{\tabcolsep}{4pt}
\renewcommand{\arraystretch}{1}
\begin{tabular}{lccc}
\toprule
sigma value & Stack-three-d1 & Hammer-cleanup-d1 & Coffee-d2 \\
\midrule
$\sigma=0$ & 82 & 78 & 70 \\
$\sigma=2$ & \textbf{90} & \textbf{88} & \textbf{78} \\
$\sigma=4$ & 86 & 78 & 72 \\
\bottomrule
\end{tabular}
\caption{Effect of soft-label width $\sigma$ on success rate (\%).}
\label{tab:sigma_ablation}
\end{wraptable}
We also conduct a parameter study on the soft-label width $\sigma$; results are reported in Table~\ref{tab:sigma_ablation}.
The width $\sigma$ controls how sharply the target probability mass is concentrated around the projected keypoint. We find that an intermediate value strikes the best balance. When $\sigma=0$, the label approaches a one-hot target with all mass on a single pixel: the supervision signal is sparse and the loss is dominated by exact-pixel matches, which provides little gradient for nearby predictions and makes training sensitive to small projection errors. When $\sigma=4$, the label becomes overly diffuse and spreads mass across a large neighborhood, blurring the target and reducing localization precision, which in turn degrades the accuracy of the triangulated 3D keypoints. A moderate $\sigma=2$ retains a well-localized peak while still providing a smooth gradient in the surrounding region, yielding the highest success rate across all three tasks.

  \subsection{Pseudocode}
  \label{appendix:pseudocode}

  Algorithms~\ref{alg:amp-train} and~\ref{alg:amp-infer} summarize the training
  and inference of \ours{} in the notation of Section~\ref{sec:method}. The
  X-Net predictor $\psi_\theta$ and the action extractor $f$ compose into the
  policy $\pi=f\circ\psi$. $\textsc{Pose2Kp}$ maps a pose to its $m{=}5$
  keypoints $p^{1:m}{:=}(p^1,\dots,p^m)$ (Fig.~\ref{fig:keypoint_repr}) and $f$
  is its geometric inverse; $\mathcal{P}_k$ projects a 3D point into side view
  $k$, and $\mathcal{T}$ triangulates the matched pixels from the $n{=}2$ side
  views back to 3D. Soft labels $\mathbf{h}_{ijk}$ are $\sigma{=}2$ Gaussians and
  a chunk spans $l$ steps. For brevity each line is written once but applies over
  all timesteps $i{\le}l$, keypoints $j{\le}m$, and side views $k{\le}n$. In
  training, lines~1--4 turn the action chunk into heatmap labels while lines~5--7
  predict the heatmaps and update $\theta$; $\textsc{CrossEntropy}$ is the
  pixel-wise loss of Section~\ref{sec:method}.

  \vspace{4pt}
  \noindent
  \begin{minipage}[t]{0.48\textwidth}
  \begin{algorithm}[H]
  \footnotesize
  \caption{\ours{} -- Training step}
  \label{alg:amp-train}
  \begin{algorithmic}[1]
  \Require demo $(\mathcal{O},\mathcal{A})$;\ $\psi_\theta$;\ rate $\eta$
  \State $p_i^{1:m}\gets \textsc{Pose2Kp}(a_i)$
  \State $(u,v)_{ijk}\gets \mathcal{P}_k(p_i^j)$
  \State jointly augment $\mathcal{O}$ and $(u,v)_{ijk}$
  \State $\mathbf{h}_{ijk}\gets \mathcal{N}\!\big((u,v)_{ijk},\sigma^2\big)$
  \State $\hat{\mathbf{h}}_{ijk}\gets \mathrm{softmax}\,\psi_\theta(\mathcal{O})$
  \State $\mathcal{L}_{\mathrm{CE}}\gets \textsc{CrossEntropy}(\hat{\mathbf{h}}_{ijk},\mathbf{h}_{ijk})$
  \State $\theta\gets\theta-\eta\,\nabla_\theta\mathcal{L}_{\mathrm{CE}}$
  \end{algorithmic}
  \end{algorithm}
  \end{minipage}
  \hfill
  \begin{minipage}[t]{0.48\textwidth}
  \begin{algorithm}[H]
  \footnotesize
  \caption{\ours{} -- Inference\ ($\pi=f\circ\psi$)}
  \label{alg:amp-infer}
  \begin{algorithmic}[1]
  \Require obs $\mathcal{O}$;\ $\psi_\theta$, $\mathcal{T}$, $f$
  \State $\hat{\mathbf{h}}_{ijk}\gets \mathrm{softmax}\,\psi_\theta(\mathcal{O})$
  \State $(u,v)_{ijk}\gets \arg\max_{(u,v)}\hat{\mathbf{h}}_{ijk}$
  \State $p_i^j\gets \mathcal{T}\big(\{(u,v)_{ijk}\}_{k=1}^{n}\big)$
  \State $a_i{=}(T_i,R_i,w_i)\gets f(p_i^{1:m})$
  \State \Return $\mathcal{A}=(a_1,\dots,a_l)$
  \State execute first $l_e$ steps, then re-plan
  \end{algorithmic}
  \end{algorithm}
  \end{minipage}

\subsection{Equivariant Data Augmentation for Heatmap Prediction}
\label{equiv_data_aug}
Representing actions as heatmaps in observation space preserves the task’s geometric symmetries.
Specifically, rotations and translations applied to the input images induce corresponding transformations in the predicted heatmaps. Moreover, transformations applied independently to each view result in independent transformations of the corresponding heatmaps without affecting others. This property corresponds to the $n$-equivariance, which we extend to heatmap prediction under a classification formulation:
\begin{equation}
\psi(GO) = G\,\psi(O),
\end{equation}
where $G = \operatorname{diag}(g_1, g_2, g_3)$ with $g_i \in SE(2)$ denotes independent transformations applied to each observation, and $O = \operatorname{diag}(o_1, o_2, o_3)$ denotes the stacked multi-view observations.

\begin{wraptable}[5]{r}{0.53\linewidth}
\vspace{-0.55em}
\centering
\scriptsize
\setlength{\tabcolsep}{2pt}
\renewcommand{\arraystretch}{1}
\resizebox{\linewidth}{!}{
\begin{tabular}{l c c c}
\toprule
Method & stack-three-d1 & hammer-cleanup-d1 & coffee-d2 \\
\midrule
Action Map Policy & 90 & 88 & 78 \\
w/o equiv. data aug. & 58 ($\downarrow$32)& 62 ($\downarrow$26) & 66 ($\downarrow$12) \\
\bottomrule
\end{tabular}
}
\caption{Ablation study on equivariant data augmentation.}
\label{tab:ablation_euqiv_aug}
\end{wraptable}
We apply equivariant data augmentation to realize the symmetries by jointly transforming each observation and its corresponding ground-truth heatmaps with random rotations and translations. This encourages the policy to learn symmetry-consistent mappings, enlarges the support of the data distribution, and biases the model toward responding to local image features, as it must steer the end effector to manipulate objects under varying transformations. We further conduct an ablation study by removing equivariant data augmentation, with the results reported in Table~\ref{tab:ablation_euqiv_aug}.

\subsection{Real-World Heatmap-Prediction Visualization}
\label{appendix:heatmap_vis}

\begin{figure*}[ht]
    \centering
    \setlength{\tabcolsep}{1pt}
    \renewcommand{\arraystretch}{0.4}

    \resizebox{\textwidth}{!}{%
    \begin{tabular}{c@{\hspace{3pt}}cccccccc}
        &
        \scriptsize Trajectory &
        \scriptsize Heatmap &
        \scriptsize Raw logits &
        \scriptsize $\log\mathrm{softmax}$ &
        \scriptsize Softmax &
        \scriptsize Target $\mathcal{N}$ &
        \scriptsize Argmax &
        \scriptsize Model trail \\

        \rotatebox{90}{\scriptsize Side view 0} &
        \includegraphics[width=0.115\textwidth]{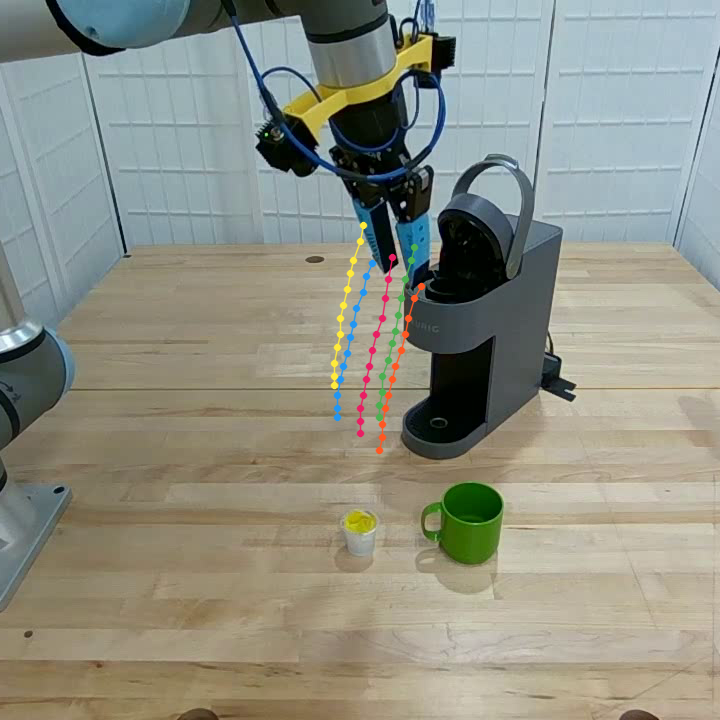} &
        \includegraphics[width=0.115\textwidth]{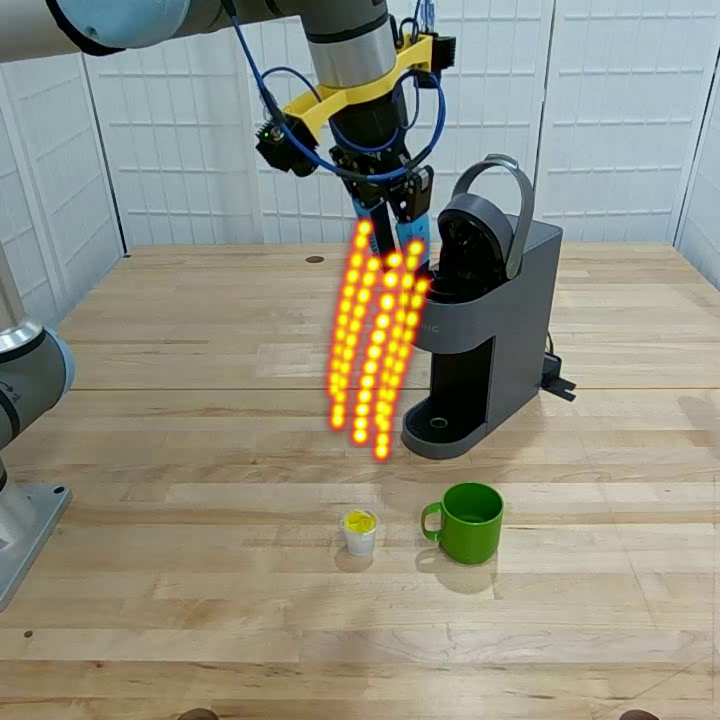} &
        \includegraphics[width=0.115\textwidth]{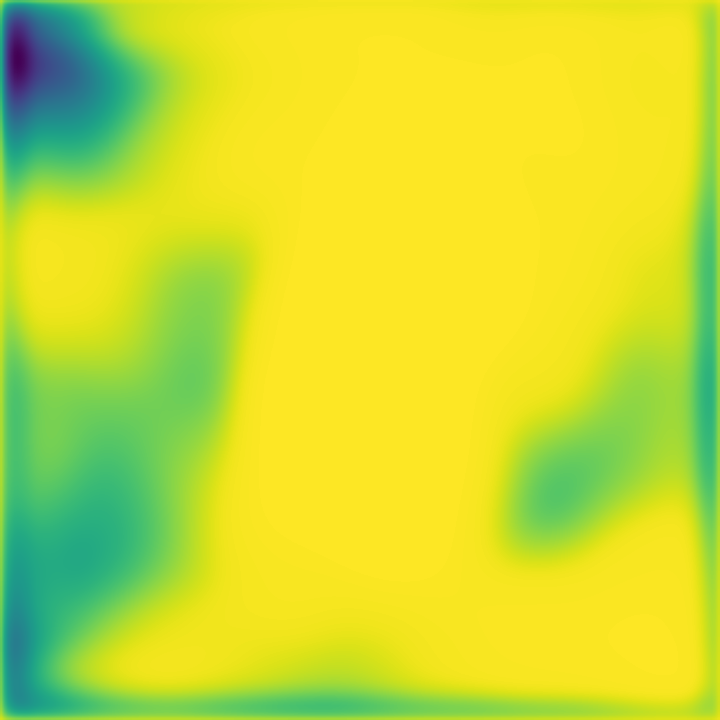} &
        \includegraphics[width=0.115\textwidth]{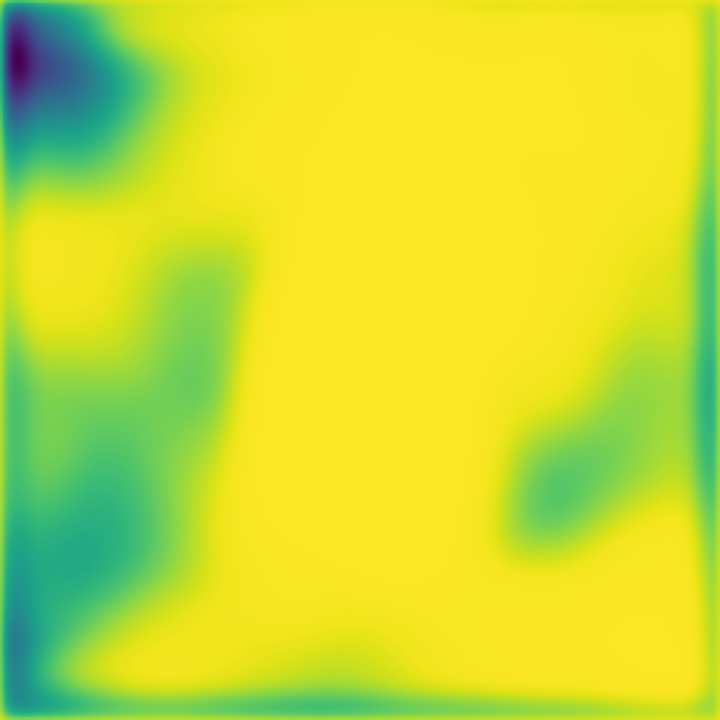} &
        \includegraphics[width=0.115\textwidth]{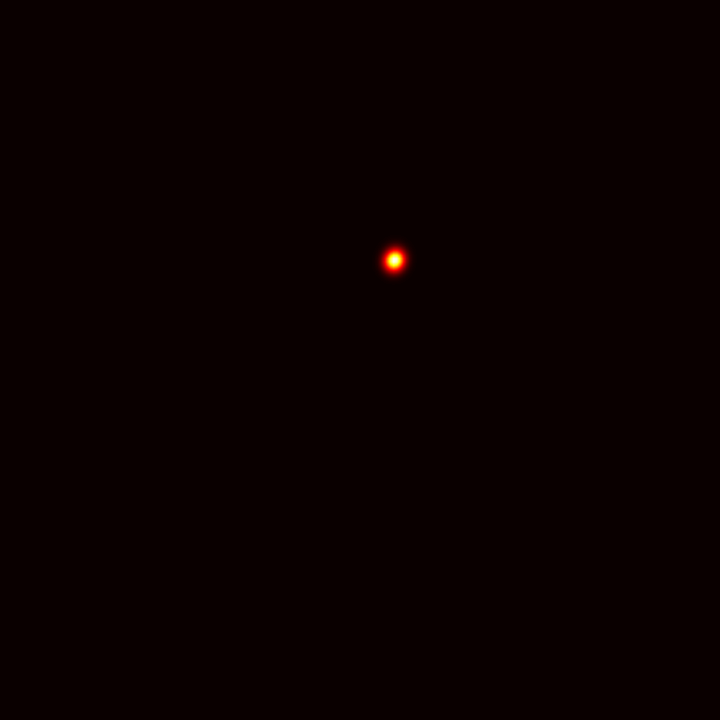} &
        \includegraphics[width=0.115\textwidth]{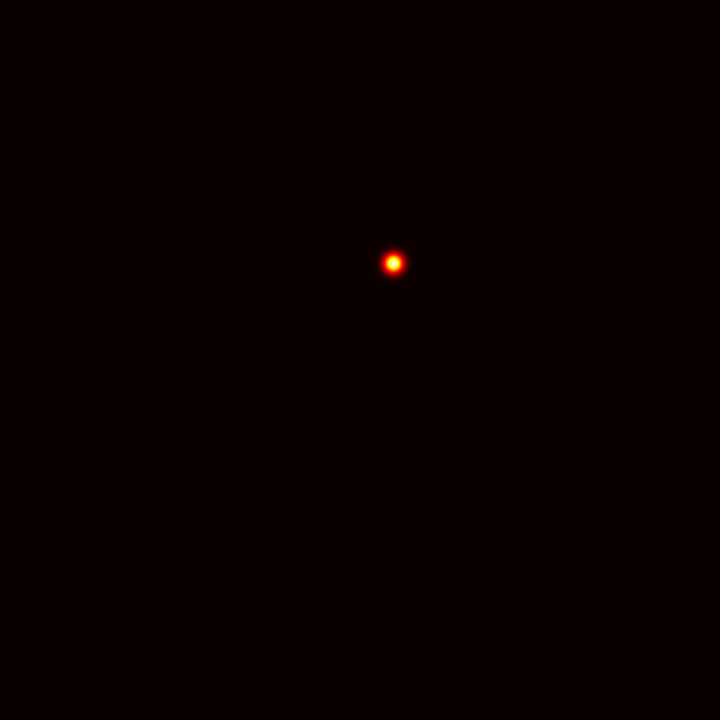} &
        \includegraphics[width=0.115\textwidth]{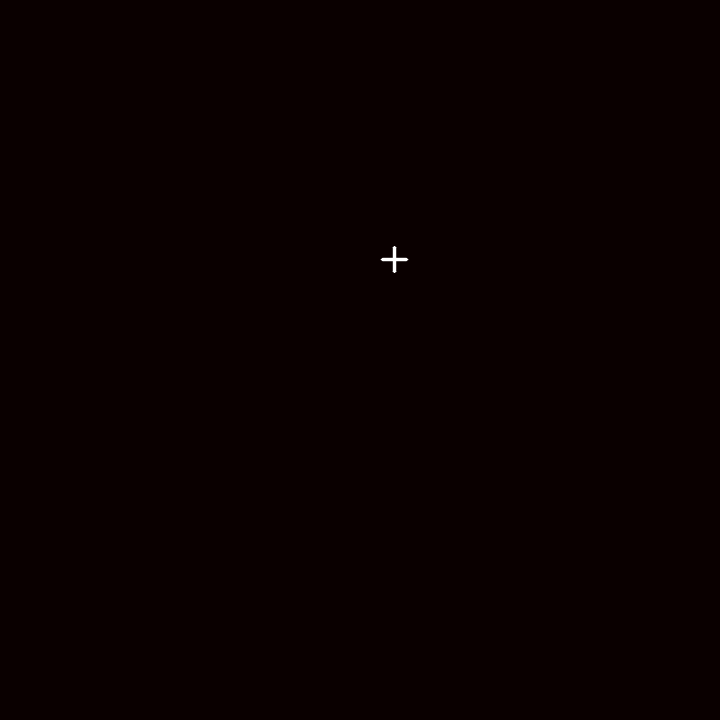} &
        \includegraphics[width=0.115\textwidth]{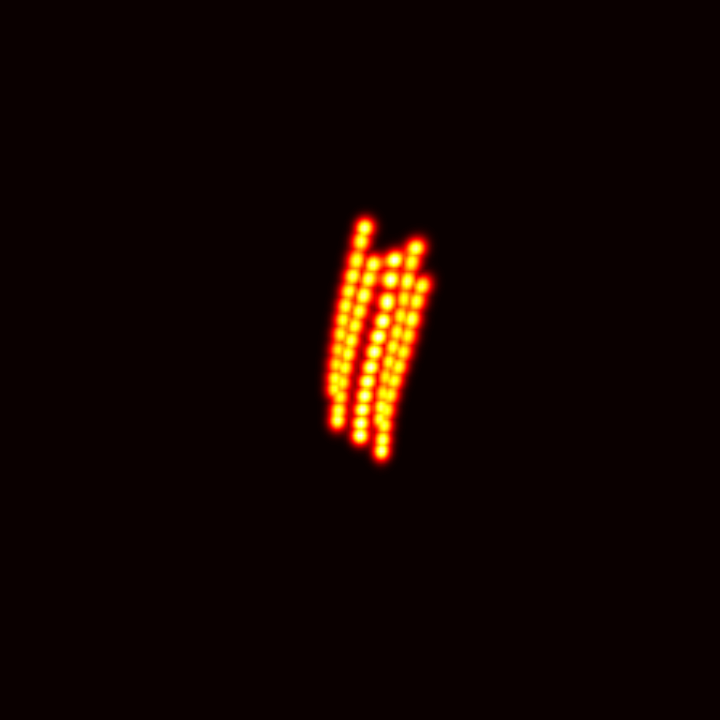} \\

        \rotatebox{90}{\scriptsize Side view 1} &
        \includegraphics[width=0.115\textwidth]{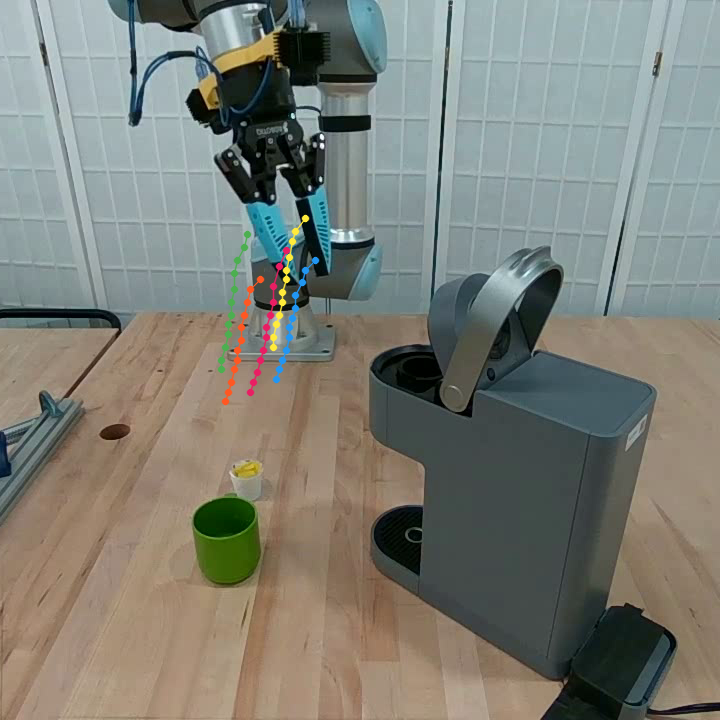} &
        \includegraphics[width=0.115\textwidth]{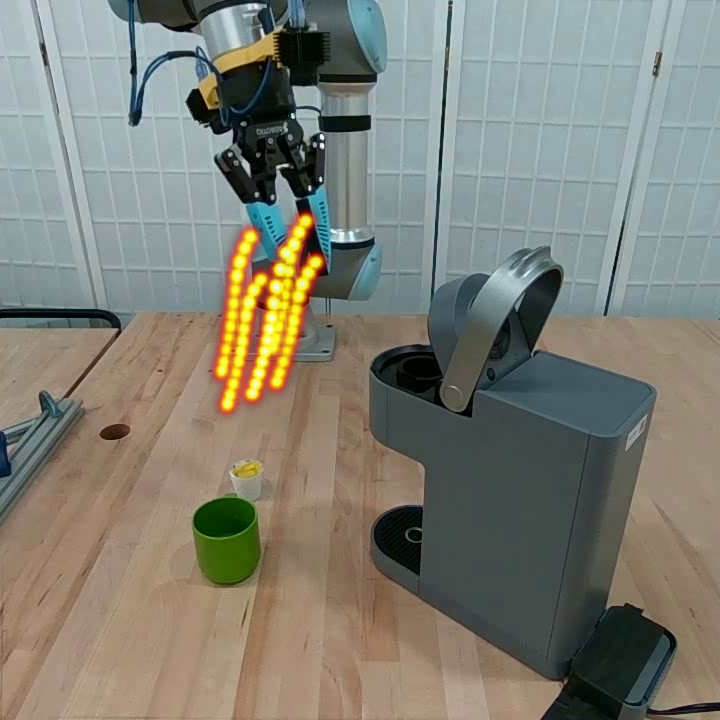} &
        \includegraphics[width=0.115\textwidth]{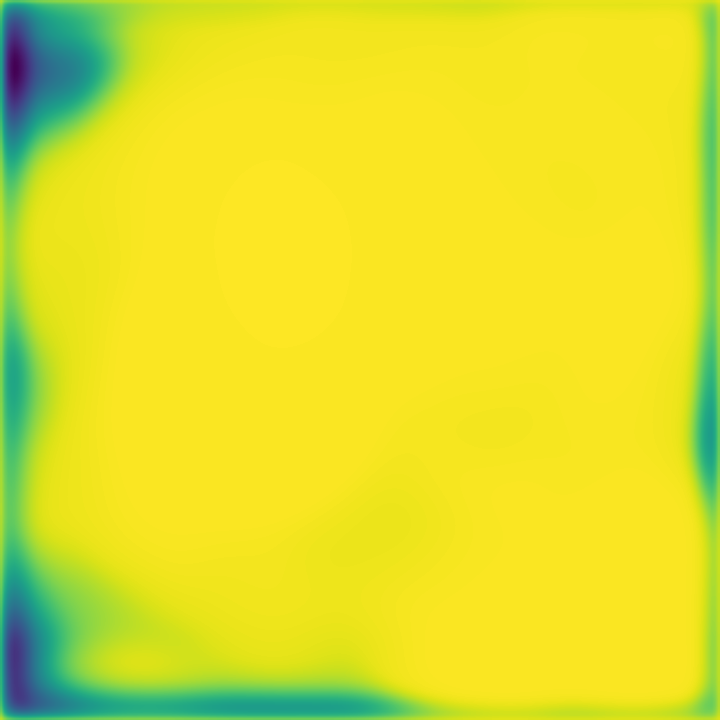} &
        \includegraphics[width=0.115\textwidth]{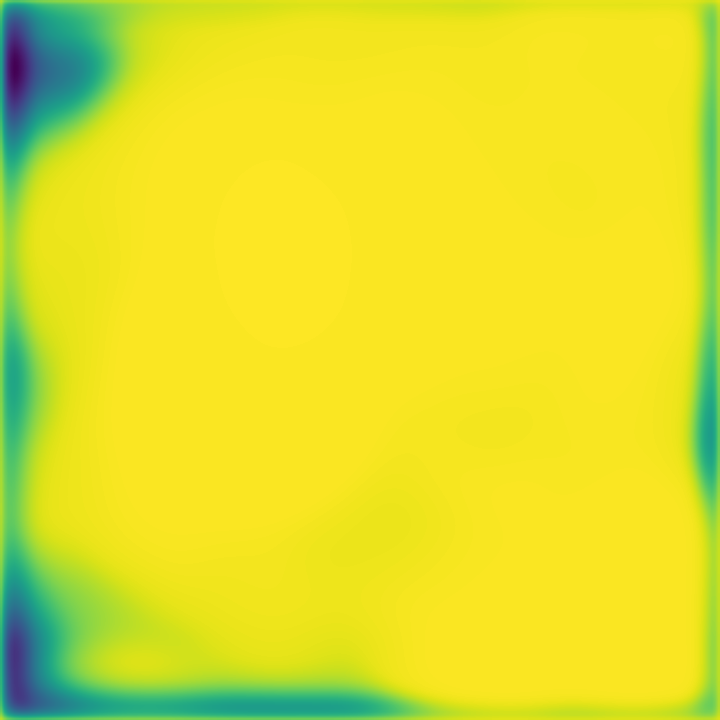} &
        \includegraphics[width=0.115\textwidth]{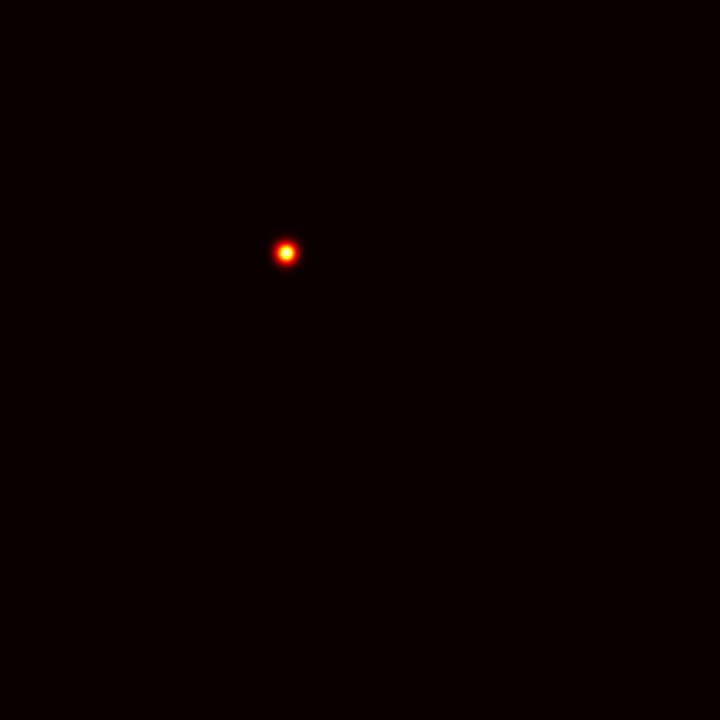} &
        \includegraphics[width=0.115\textwidth]{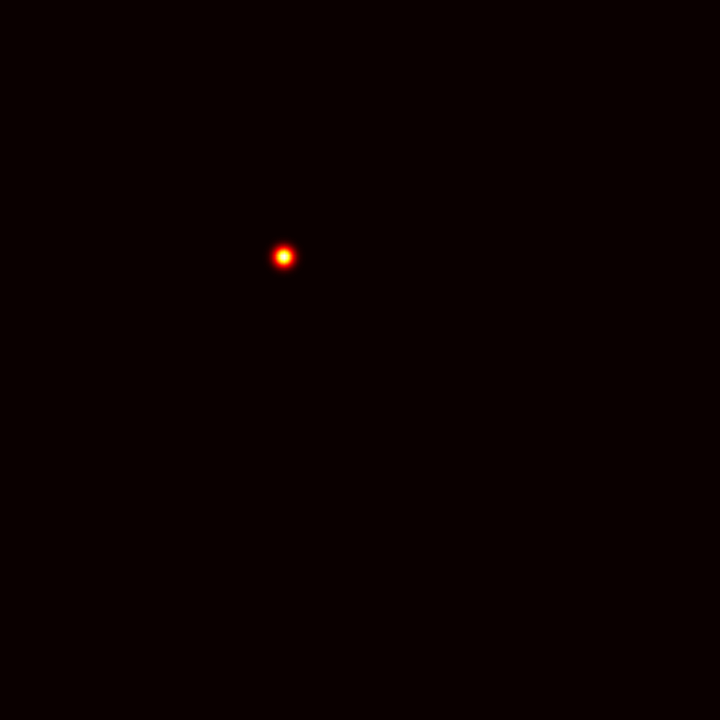} &
        \includegraphics[width=0.115\textwidth]{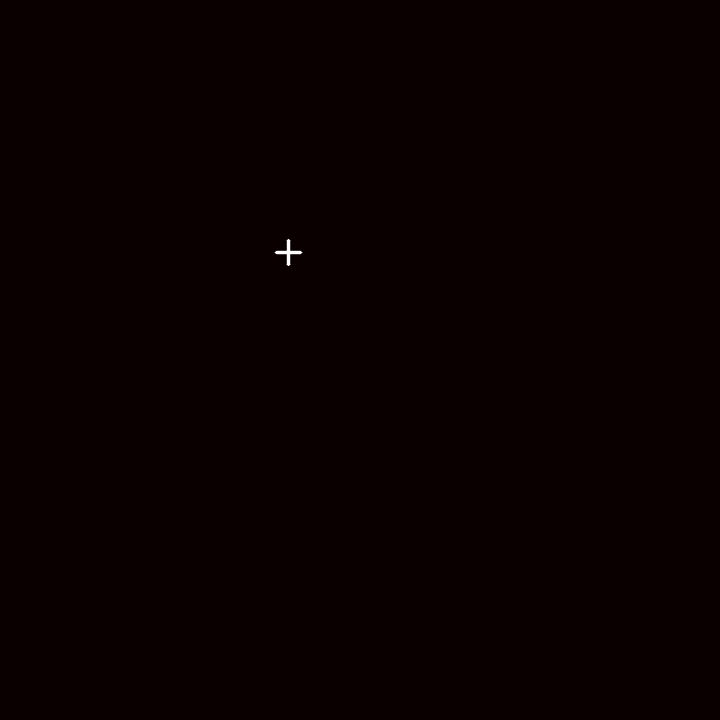} &
        \includegraphics[width=0.115\textwidth]{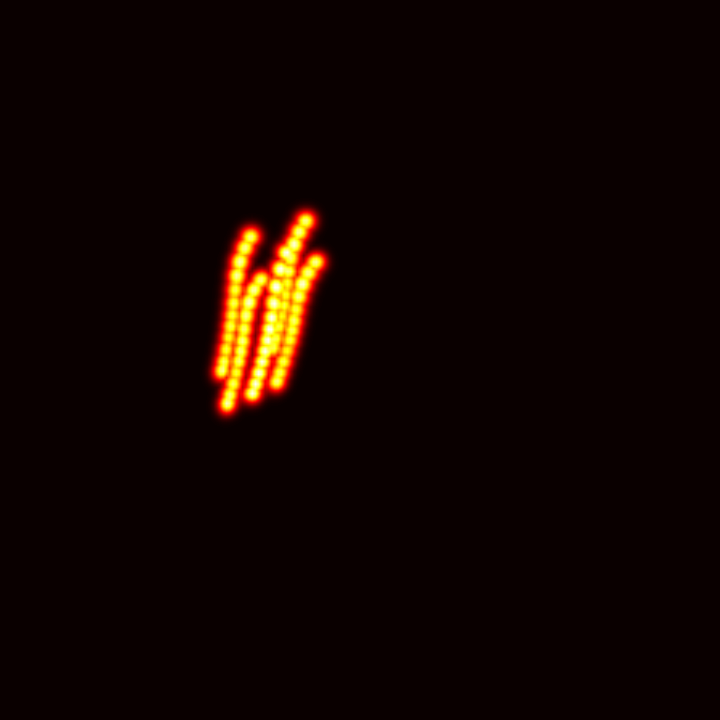} \\
    \end{tabular}
    }

    \caption{Heatmap prediction visualization on the \textsc{Make-Coffee} task, evaluated at the midpoint of a downward grasp toward the coffee pod. Each row corresponds to one calibrated side view. The columns show the dense prediction process, including the raw decoder output, cross-entropy supervision, spatial softmax, argmax pixel selection, and the final horizon-wide trajectory. Columns (1)--(2) summarize the full $5{\times}12$ keypoint-step output of the policy, while Columns (3)--(7) visualize the image-space distribution of a single keypoint.}
    \label{fig:coffee_pipeline}
\end{figure*}
The X-Net casts trajectory prediction as a dense per-pixel classification
problem. For every keypoint $k\!\in\!\{1,\dots,K\}$ and every future
horizon step $h\!\in\!\{1,\dots,H\}$, it outputs a single
$224\!\times\!224$ score map for each side view. We do not apply any
non-linearity to this map before supervision. During training, the raw
logits are optimized with a spatial cross-entropy loss against a Gaussian
target. During inference, we take the $\operatorname{arg\,max}$ pixel and
lift the predicted pixels from two calibrated views into a 3D keypoint
cloud via triangulation.

To make this pipeline transparent, we visualize each intermediate
quantity for the \textsc{Make-Coffee} task on a single frame captured midway
through a downward grasping motion. From left to right in
Fig.~\ref{fig:coffee_pipeline}, the eight columns are:
\begin{itemize}
    \setlength{\itemsep}{1pt}
    \item \textbf{(1) Trajectory.} Per-keypoint $\operatorname{arg\,max}$
          pixels over the full $H{=}12$ prediction horizon, connected as
          polylines and overlaid on the native-resolution side-view
          image. This is the 2D output the policy commits to
          \emph{before} triangulation.
    \item \textbf{(2) Heatmap.} The same prediction in the heatmap
          domain: the per-step spatial-softmax maps are summed over the
          $5$ keypoints, peak-normalized, max-merged across the horizon,
          and alpha-blended onto the native background. The bright
          ``trail'' is the heatmap analogue of the trajectory in
          column (1).
    \item \textbf{(3) Raw logits.} The unactivated output of the X-Net
          decoder, $\mathrm{logit}_{i,j}\!\in\!\mathbb{R}$, rendered by
          linearly stretching its values to a colormap.
    \item \textbf{(4) Log-softmax.} The quantity used by the
          cross-entropy loss, obtained by applying $\log\mathrm{softmax}$
          over all image pixels. Its range is $(-\infty,0]$, with $0$ at
          the peak. Since $\log\mathrm{softmax}$ differs from the raw
          logits only by an additive normalization term, its spatial
          pattern matches column (3).
    \item \textbf{(5) Spatial softmax.} Exponentiating the logits and
          normalizing them to unit mass over all pixels converts the
          raw score map into a probability distribution, often appearing
          as a sharply peaked heatmap.
    \item \textbf{(6) Training target.} The supervision used during
          training: a $\sigma{=}2$ Gaussian centered at the ground-truth
          keypoint pixel and normalized to unit mass.
    \item \textbf{(7) Argmax.} The single highest-scoring pixel, which is
          the only information passed downstream to two-view
          triangulation.
    \item \textbf{(8) Model trail.} A horizon-wide rendering of the
          policy prediction, obtained by stacking the spatial-softmax
          maps for all $5$ keypoints and $H{=}12$ steps,
          peak-normalizing each step, and taking the maximum across
          steps.
\end{itemize}

Two observations follow. First, columns (3) and (5) encode the same
ordering over pixels, but their visual appearance can be very different:
spatial softmax can turn a nearly flat logit map into a compact,
Gaussian-like blob. Second, the model prediction in column (5) closely
matches the training target in column (6), suggesting that the
cross-entropy objective has converged well in the heatmap domain.

\subsection{Real-world Experiment Details}
In the real-world experiments, we apply color jittering and use exactly the same observations for our method and all baselines to ensure a fair comparison. The results in Table \ref{tab:realworld_long_horizon} show that our method significantly outperforms the baselines across all tasks. In addition, our method achieves an inference speed of 13.80 ms. This latency is measured from receiving the observation to outputting the full trajectory commands, rather than only the model forward-pass time. All inference times are measured on a single NVIDIA RTX 3090 GPU. We note that the SVD-based triangulation step contributes non-negligible overhead, and can be further optimized.

\subsection{Real-World Experiments: Failure Case Analysis}
We report the completion rate without partial credit for each task. The main failure mode of our method is imprecise grasping and insertion, occurring primarily under out-of-distribution evaluation. These failures are more severe for the diffusion-policy and ACT baselines, which commonly either get stuck at a fixed location or collapse toward the averaged position of several candidate objects. In contrast, AMP rarely exhibits this collapse, as its classification-based formulation preserves the multi-modal distribution over candidate grasps.

\end{document}